\documentclass[lettersize,journal]{IEEEtran}
\usepackage{amsmath,amsfonts}
\usepackage{algorithmic}
\usepackage{algorithm}
\usepackage{array}
\usepackage{textcomp}
\usepackage{stfloats}
\usepackage{url}
\usepackage{verbatim}
\usepackage{graphicx}
\usepackage{cite}
\usepackage{xcolor}
\usepackage{multirow}
\usepackage{multicol}
\usepackage{microtype}
\usepackage{subfigure}
\usepackage{booktabs}
\usepackage{makecell}
\usepackage{float}
\usepackage[breaklinks=true,bookmarks=false]{hyperref}

\begin{document}

\title{A Spectral Perspective towards Understanding and Improving Adversarial Robustness}

\author{Binxiao~Huang, Rui~Lin, Chaofan~Tao, and ~Ngai Wong,~\IEEEmembership{Senior Member,~IEEE}}

%\author{IEEE Publication Technology,~\IEEEmembership{Staff,~IEEE,}
        % <-this % stops a space
%\thanks{This paper was produced by the IEEE Publication Technology Group. They are in Piscataway, NJ.}% <-this % stops a space
%\thanks{Manuscript received April 19, 2021; revised August 16, 2021.}}

\markboth{Journal of \LaTeX\ Class Files,~Vol.~14, No.~8, August~2021}%
{Shell \MakeLowercase{\textit{et al.}}: A Sample Article Using IEEEtran.cls for IEEE Journals}


\maketitle

\begin{abstract}
Deep neural networks (DNNs) are incredibly vulnerable to crafted, imperceptible adversarial perturbations. While adversarial training (AT) has proven to be an effective defense approach, the AT mechanism for robustness improvement is not fully understood. This work investigates AT from a spectral perspective, adding new insights to the design of effective defenses. In particular, we show that AT induces the deep model to focus more on the low-frequency region, which retains the shape-biased representations, to gain robustness. Further, we find that the spectrum of a white-box attack is primarily distributed in regions the model focuses on, and the perturbation attacks the spectral bands where the model is vulnerable. Based on this observation, to train a model tolerant to frequency-varying perturbation, we propose a spectral alignment regularization (SAR) such that the spectral output inferred by an attacked adversarial input stays as close as possible to its natural input counterpart. Experiments demonstrate that SAR and its weight averaging (WA) extension could significantly improve the robust accuracy by 1.14\% $\sim$ 3.87\% relative to the standard AT, across multiple datasets (CIFAR-10, CIFAR-100 and Tiny ImageNet), and various attacks (PGD, C\&W and Autoattack), without any extra data.
\end{abstract}


%%%%%%%%% BODY TEXT
\section{Introduction}
Deep Neural Networks (DNNs) have exhibited strong capabilities in various application such as computer vision~\cite{he2016deep}, natural language processing~\cite{devlin2018bert}, etc. 
However, researches show that even well-trained DNNs are highly susceptible to adversarial perturbations~\cite{goodfellow2014explaining,szegedy2013intriguing}. These perturbations are nearly indistinguishable to human eyes but can mislead neural networks to completely erroneous outputs, thus endangering safety-critical applications and hindering practical deployments. 

%To evaluate the adversarial robustness of the well-trained models, 
Various attacks (e.g., Projected Gradient Descent (PGD)~\cite{madry2017towards}, AutoAttack (AA)~\cite{zhang2020attacks}) have been proposed, which pose a huge challenge to defense methods.
% Projected gradient descent (PGD)~\cite{madry2017towards} attack is proved as one of the strongest first-order attacks and is commonly used to generate adversarial inputs to verify robustness.
Among defense methods for the true robustness~\cite{das2018shield, mao2019metric, zheng2020efficient}, 
adversarial training (AT)~\cite{ madry2017towards}, which feeds adversarial inputs into a DNN to solve a min-max optimization problem, proves to be an effective means without obfuscated gradients  ~\cite{athalye2018obfuscated}. Based on AT, some works~\cite{wong2020fast, sriramanan2020guided, andriushchenko2020understanding, sriramanan2021towards, jia2022prior} focus on accelerating the training with simple attacks, others~\cite{rice2020overfitting,zhang2019theoretically,wu2020adversarial,jia2022adversarial} take advantage of different regularizations or training strategies to improve the adversarial robustness.

On the other hand, frequency analysis provides a new lens on the generalization behavior of DNNs. Wang \textit{et al.}~\cite{wang2020high} claim that convolutional neural networks (CNNs) have the ability to capture human-imperceptible high-frequency components of images for predictions. Xie \textit{et al.}~\cite{xie2021learning} propose a frequency-aware dynamic network for efficient super-resolution, Yin \textit{et al.}~\cite{yin2019fourier} establish a connection between the frequency of common corruptions and model performance, especially for high-frequency corruptions. These findings motivate us to zoom in deeper on AT from a spectral viewpoint. %\huang{Many papers constrain the input in the frequency domain, which will reduce the information fed into the model. Could we input the original images without filtering and place frequency constraints on the output to force the model to extract the desired information to improve the robustness automatically?}
Specifically, we obtain models with different frequency biases and study the distribution of their corresponding white-box attack perturbations across different datasets. We then propose a simple yet effective spectral alignment regularization (SAR) to improve the adversarial robustness, followed by validation on multiple datasets against various attacks. Our main contributions are:
\begin{itemize}
\item We find that AT induces a model to focus primarily on low frequencies, where the shape-biased representation resides, to improve the robustness. In contrast, solely focusing on low-frequency information \textit{does not} lead to adversarial robustness (cf. Table~\ref{lpf_accuracy} and Figure~\ref{lpf_x_prime}).

% 这里使用not only, but also是因为有文章指出了扰动是dataset dependent, 但是没有说model dependent.
% We discover that the frequency distribution of perturbation is not only related to the dataset but also to the model.
\item We reveal \emph{for the first time} that white-box attacks mainly occur in the frequencies the model focuses on, and can adjust their aggressive frequency distribution to match the model's vulnerability to frequency corruptions. This explains why white-box attacks are often hard to defend. %This is a new sight to investigate the aggressive frequency distribution of perturbations. 

\item We propose the SAR that enforces alignment of the outputs of natural and adversarial examples in the frequency domain, thus effectively boosting the adversarial robustness.
\end{itemize}

%-------------------------------------------------------------------------%

\section{Related Works}
AT has evolved to be among the de facto schemes to obtain adversarial robustness in a DNN. Some recent results inspired by AT are also in place to further boost the robust accuracy.
TRADES~\cite{zhang2019theoretically} identifies a trade-off between standard and robust accuracies that serves as a guiding principle for designing the defenses. 
Andrew \textit{et al.}~\cite{ilyas2019adversarial} argue that adversarial inputs are features instead of bugs, and the understanding of humans about the input is quite different from that of networks.
MART~\cite{wang2019improving} explicitly distinguishes the misclassified and correctly classified examples to boost the robustness.  %MART
RST~\cite{carmon2019unlabeled} achieves high robust accuracy through semi-supervised learning with extra unlabeled data.
AWP~\cite{wu2020adversarial} claims that the weight loss landscape is closely related to the robust generalization gap, and proposes an effective adversarial weight perturbation method to overcome the robust overfitting problem~\cite{rice2020overfitting}. 
LAS-AT~\cite{jia2022adversarial} introduces a learnable attack strategy to produce proper hyperparameters for generating the perturbations during training to improve the robustness. None of these methods is analyzed from the perspective of the frequency domain.

Wang \textit{et al.}~\cite{wang2020high} find that robust models have smooth convolutional kernels in the first layer, thereby paying more attention to low-frequency content. Yin \textit{et al.}~\cite{yin2019fourier} view AT as a data augmentation method to bias the model toward low frequencies, which improves the robustness to high-frequency corruptions at the cost of reduced robustness to low-frequency corruptions. This, however, could not explain why AT makes the model robust when the perturbations are concentrated in the low-frequency region in some datasets. Zhang \textit{et al.}~\cite{zhang2019interpreting} find that AT-CNNs are better at capturing long-range correlations (e.g., shapes), and less biased towards textures than naturally trained CNNs in object recognition datasets.
Wang \textit{et al.}~\cite{wang2020towards} %state that perturbations mainly focus on the high-frequency information in natural images, and low-frequency information is more robust than the high-frequency part. 
claim that developing a stronger association between low-frequency information with true labels makes the model robust. However, our study shows that building this connection alone cannot render the model adversarially robust. The closest work to ours is~\cite{maiya2021frequency}, which discovers that the adversarial perturbation is data-dependent. Our research goes one step further to show that the perturbation is also model-dependent, and explains why it behaves differently across the datasets and models. This leads to our proposed SAR to enhance robust accuracy.

%-------------------------------------------------------------------------%

\section{Preliminaries}
% \subsection{Adversarial Training}
Typically, AT updates the model weights to solve the min-max saddle point problem:
\begin{equation}
\min _{\theta} \frac{1}{n} \sum_{i=1}^{n} \max _{\left\|\delta\right\|_{p} \leq \epsilon} \mathcal{L}\left(f_{\theta}\left(\mathbf{x}_{i}+\delta\right), y_{i}\right), 
\end{equation}
where $n$ is the number of training examples, $\mathbf{x}_{i}+\delta$ is the adversarial input within the $\epsilon$-ball (bounded by an $L_p$-norm) centered at the natural input $\mathbf{x}_{i}$, $\delta$ is the perturbation, $y_{i}$ is the true label, \textcolor{black}{$f_{\theta}$ denotes the DNN with weight $\theta$, and $\mathcal{L}(\cdot)$ represents the classification loss (e.g., cross-entropy (CE))}. 

\begin{figure}[h]
\begin{center}
%\framebox[4.0in]{$\;$}
\includegraphics[width=1.0\linewidth]{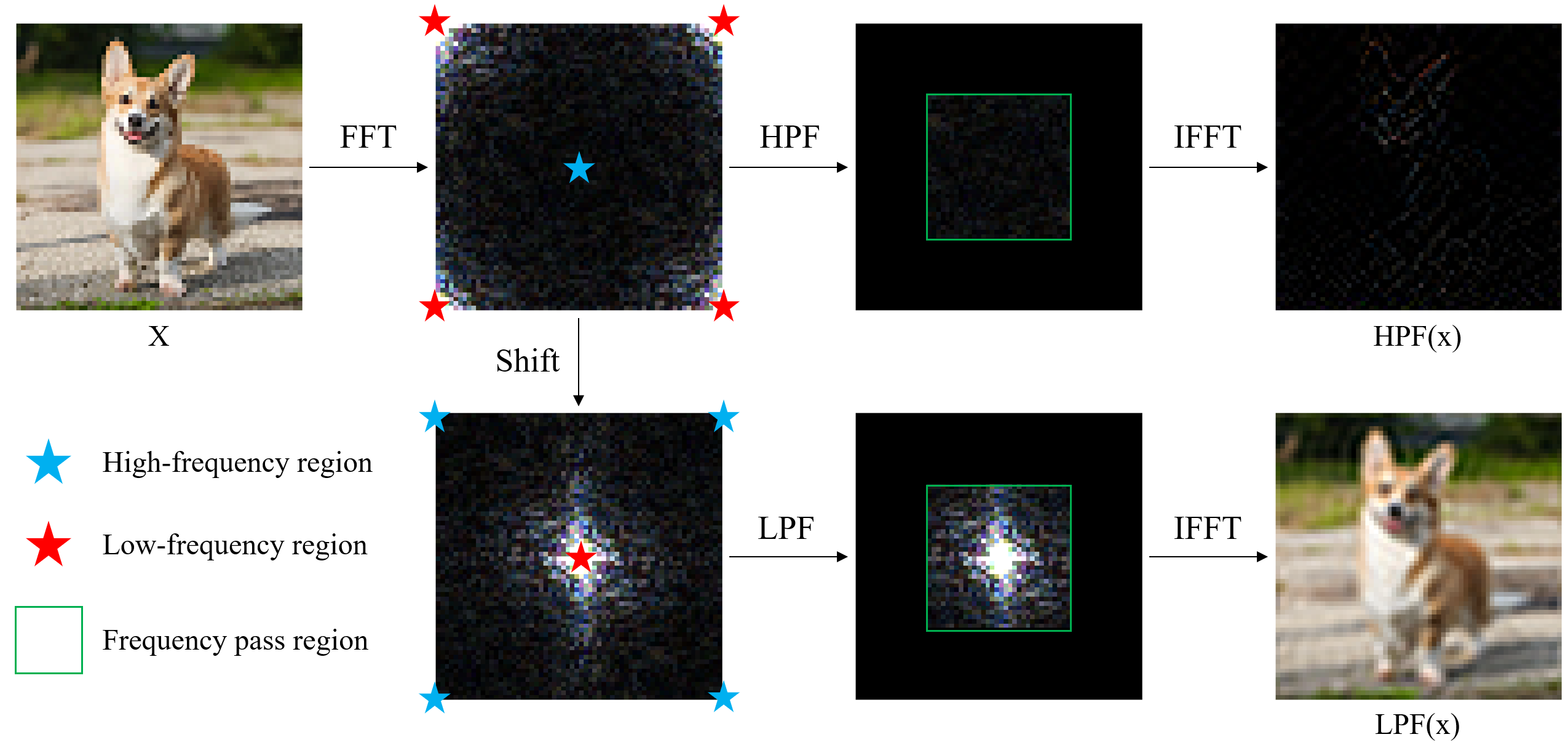}
\end{center}
\caption{HPF and LPF ($k$=32) of a (3$\times$64$\times$64) image .}
\label{hpf_lpf}
\end{figure}

We refer to the adversarially trained model as the \emph{robust model} and the naturally trained model as the \emph{natural model}. The accuracies achieved on natural and adversarial inputs are denoted as \emph{standard} and \emph{robust accuracy}, respectively. \textcolor{black}{We define the high-pass filtering (HPF) with bandwidth $k$ as an operation having three stages: 1) applying Fast Fourier Transform (FFT) on the input, 2) preserving the $k \times k$ patch in the center (viz. high frequencies) and zeroing all external values of the intermediate result, and 3) applying the inverse FFT (IFFT) to get the final output.}
Low-pass filtering (LPF) is defined similarly except \textcolor{black}{in step 2} that the low-frequency part is shifted to the center after FFT \textcolor{black}{and will} be preserved by the center $k \times k$ patch as in~\cite{yin2019fourier} and illustrated in Figure~\ref{hpf_lpf}.

\section{Empirical Observation and Analyses}
\subsection{Frequency Attention}
\label{LFIandshape}
\begin{figure}[tp]
\begin{center}
%\framebox[4.0in]{$\;$}
\includegraphics[width=1.0\linewidth]{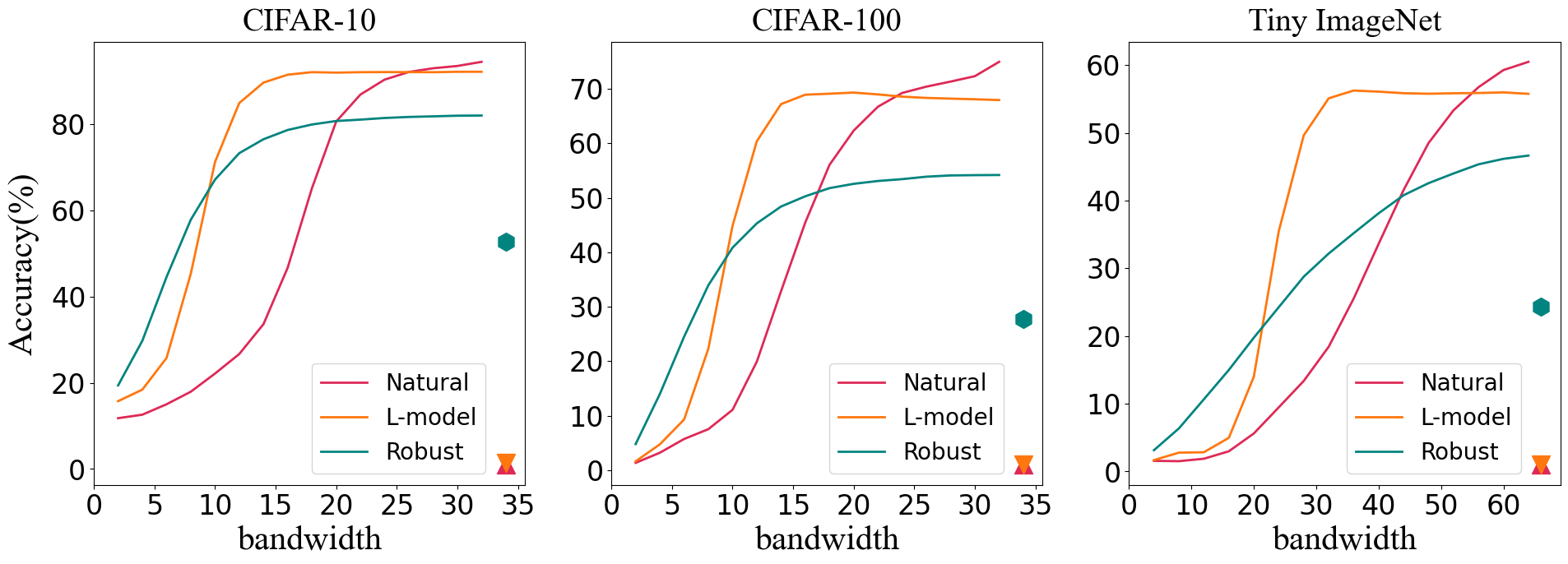}
\end{center}
\caption{Top-1 accuracy(\%) of natural, L- and robust ResNet18 models for clean inputs processed by LPF with different bandwidths (cf. Table~\ref{lpf_accuracy}). The robust accuracy against PGD-20 is indicated by same-color markers.}
\label{lpf_acc_fig}
\end{figure}

\paragraph{Attention to the Frequency Domain}  \textcolor{black}{Wang \textit{et al.}~\cite{wang2020high} claims that labels are inherently tied to low-frequency information, which indicates that standard accuracy is closely related to low-frequency information.
Based on this observation, we aim to explore the connection between low-frequency information and adversarial robustness. Except for the natural and robust models, we train models with the LPF-processed natural inputs (denoted as L-models) with a fixed bandwidth of 16 for multiple datasets (32 for Tiny ImageNet).} %For TinyImageNet, it is a special case where the bandwidth is 32.} %Since the labels are inherently tied with the low-frequency information~\cite{wang2020high}, to maintain high standard accuracy and explore the connection between the low-frequency information and adversarial robustness, except for the natural and robust models, we train models (denoted as L-models) with the natural inputs after the LPF with a fixed bandwidth of 16 for multiple datasets (32 for Tiny ImageNet). 
\textcolor{black}{Therefore, the prediction of the L-model relies only on the low-frequency information in the image.  The optimization target of a L-model is given below:}
%So the L-models can only obtain low-frequency information in the images for prediction. The optimization target of L-model is shown in Eqn.~\ref{L-equation}. 

\begin{equation}
\label{L-equation}
\min _{\theta} \frac{1}{n} \sum_{i=1}^{n} \mathcal{L}\left(f_{\theta}\left(LPF_{16}(\mathbf{x}_{i}\right), y_{i}\right), 
\end{equation}
\textcolor{black}{Next}, we feed LPF-processed natural inputs with different bandwidths into natural, L- and robust models to evaluate the accuracy, respectively. \textcolor{black}{The numbers are in Table~\ref{lpf_accuracy} whose trends are visualized in Figure~\ref{lpf_acc_fig}, wherein the accuracy reflects how much attention the models pay to low-frequency information.} %which reflects how much attention the models pay to low-frequency information. Results are shown in Table~\ref{lpf_accuracy}. 

% The robust accuracy of the L-model is almost zero in the face of the PGD-20 attack, suggesting that learning the low-frequency information alone does not contribute to robust accuracy.

For natural models, the standard accuracy gradually increases as the bandwidth increases, indicating that the models utilize both low- and high-frequency information, \textcolor{black}{which is} consistent with the findings of~\cite{wang2020high}. %\textcolor{black}{However, } SVHN is an exception, \textcolor{black}{in which the information is mainly concentrated in low-frequency region~\cite{bernhard2021impact}.}  %as the information in this dataset . So the models trained on SVHN all primarily rely on the low-frequency information for predictions. 
For L-models, when the bandwidth increases to beyond 16 (32 for Tiny ImageNet), the standard accuracy no longer improves significantly, which is in line with the expectation that a model trained with low-frequency information focuses mainly on the low-frequency region for predictions. As for robust models, even though a large amount of high-frequency information is removed, there is only a negligible loss in standard accuracy on the CIFAR datasets. For Tiny ImageNet, the high-frequency content can further improve the standard accuracy a little, but the accuracy improvement relies mainly on low-frequency part. 

Comparing the natural, L- and robust models across different datasets leads to the conclusion that AT enforces the AT-trained model to focus primarily on low-frequency information. Furthermore, taking the CIFAR-10 as an example, the standard accuracy (81.98\%) of the robust model is similar to that of the natural model at a LPF bandwidth of 20 (80.58\%). Such observation indicates that the low standard accuracy of natural dataset in the robust model is due to the under-utilization of high-frequency components.

\begin{table*}[ht]
\centering
\caption{Top-1 accuracy(\%) of natural, L- and robust ResNet18 models subject to clean input, except the PGD-20 column which shows the robust accuracy against PGD-20 attack. The bandwidth row denotes the LPF bandwidth ($k$) applied to the inputs. The higher the $k$ the more information is retained (i.e., 32 in CIFAR or 64 in Tiny ImageNet means no filtering). Bold numbers indicate the best.}
\label{lpf_accuracy}
\begin{center}
\resizebox{\textwidth}{!}{
\begin{tabular}{c|c|cccccccc|c}
\hline
%\multicolumn{1}{c}{Dataset}  & \multicolumn{1}{c}{PGD-20}
Dataset & Bandwidth & 4 & 8 & 12 & 16 & 20 & 24 & 28 & 32 & PGD-20
\\ \hline 
% \multirow{3}{*}{SVHN} 
% & Natural & 49.07 & 83.87 & 94.82 & 95.41 & 95.50 & 95.53 & \textbf{95.60} & 95.58 & 0.38 \\
% & L-model & 49.29 & 90.11 & 94.99 & 95.64 & 95.69 & \textbf{95.71} & 95.70 & 95.69 & 0.41\\
% & Robust  & 40.47 & 81.54 & 88.43 & 89.30 & 89.36 & 89.38 & 89.38 & \textbf{89.39}  & \textbf{53.74}\\
% \hline

\multirow{3}{*}{CIFAR-10}  
& Natural & 12.60  & 17.93 & 26.67 & 46.71 & 80.58 & 90.31 & 92.94 & {94.43} & 0.0\\
& L-model & 18.41 & 45.29 & 84.87 & 91.45 & 92.02 & {92.06} & 92.03 & 92.03  & 0.05\\
& Robust  & 29.74 & 57.82 & 73.25 & 78.63 & 80.71 & 81.41 & 81.79 & {81.98}  & {51.69}\\

\hline 
\multirow{3}{*}{CIFAR-100} 
& Natural &  3.20 &  7.52 & 19.91 & 45.46 & 62.34 & 69.24 & 71.33 & {74.95} & 0.0 \\
& L-model &  4.74 & 22.29 & 60.38 & 68.92 & {69.31} & 68.56 & 68.20 & 67.94 & 0.0 \\
& Robust  & 13.99 & 33.94 & 45.30 & 50.27 & 52.56 & 53.42 & 54.1  & {54.18} & {27.81}\\
\hline 
\hline
Dataset & Bandwidth & 8 & 16 & 24 & 32 & 40 & 48 & 56 & 64 & PGD-20\\

\hline
\multirow{3}{*}{Tiny ImageNet} 
& Natural &  1.52 &  2.98 &  9.47 & 18.42 & 33.61 & 48.50 & 56.74 & {60.48} & 0.0\\
& L-model &  2.78 &  4.97 & 35.48 & {56.80} & 56.28 & 55.77 & 55.88 & 55.76 & 0.0\\
& Robust  &  6.36 & 14.99 & 24.26 & 32.18 & 38.12 & 42.57 & 45.35 & {46.64} & {23.33}\\
\hline 

\end{tabular}}
\end{center}
\end{table*}

\paragraph{Robust Features in the Low-frequency Region} Although both the L-model and robust model rely on low-frequency information for predictions, the L-model has \textit{no resistance} to PGD-20 (cf. the last column in Table~\ref{lpf_accuracy}), suggesting that learning the low-frequency information alone \textit{does not} instill robust accuracy. Besides, when the input retains limited low-frequency information, the robust model is more accurate than the natural model, and even more accurate than the L-model in particularly small bandwidth (e.g., 4, 8). This implies the robust model can extract more useful information from the particularly low-frequency region. To explore what robust information the model is concerned with, we visualize the natural images, the perturbed images, and the LPF-processed images from the CIFAR-10 dataset in Figure~\ref{lpf_x_prime}. Other datasets show similar performance, as depicted in Appendix~\ref{app1}. When $k=8$, the robust model achieves a much higher standard accuracy (57.82\%) than the natural model (17.93\%) with very limited low-frequency information on CIFAR-10.  Compared to natural images, the filtered images retain the outer contours, but the detailed textures are heavily blurred. And for the perturbed images, the texture of the foreground and background is disturbed, while the shapes are almost unaffected. Indeed, Geirhos \textit{et al.}~\cite{geirhos2018imagenet} show that naturally trained CNNs are strongly biased toward recognizing textures rather than shapes, which accounts for the low standard accuracy for the filtered images and their vulnerability. The robust model can maintain a certain level of standard and robust accuracies, when the texture is heavily smoothed whereas the shape profile is partially preserved. This indicates that AT enables the model to learn a shape-biased representation that is more human-like and consistent with the finding of~\cite{zhang2019interpreting}. Consequently, AT induces model to primarily focus on low-frequency information to learn a more shape-biased representation, which gains the robustness. % namely, shape-based features improves robustness, and low-frequency information preserves objects' shapes while blurring textures. Consequently, AT induces robustness in models. to primarily focus on low-frequency information to learn a more shape-biased representation, which gains the robustness. 

\begin{figure}[h]
\begin{center}
%\framebox[4.0in]{$\;$}
\includegraphics[width=1.0\linewidth]{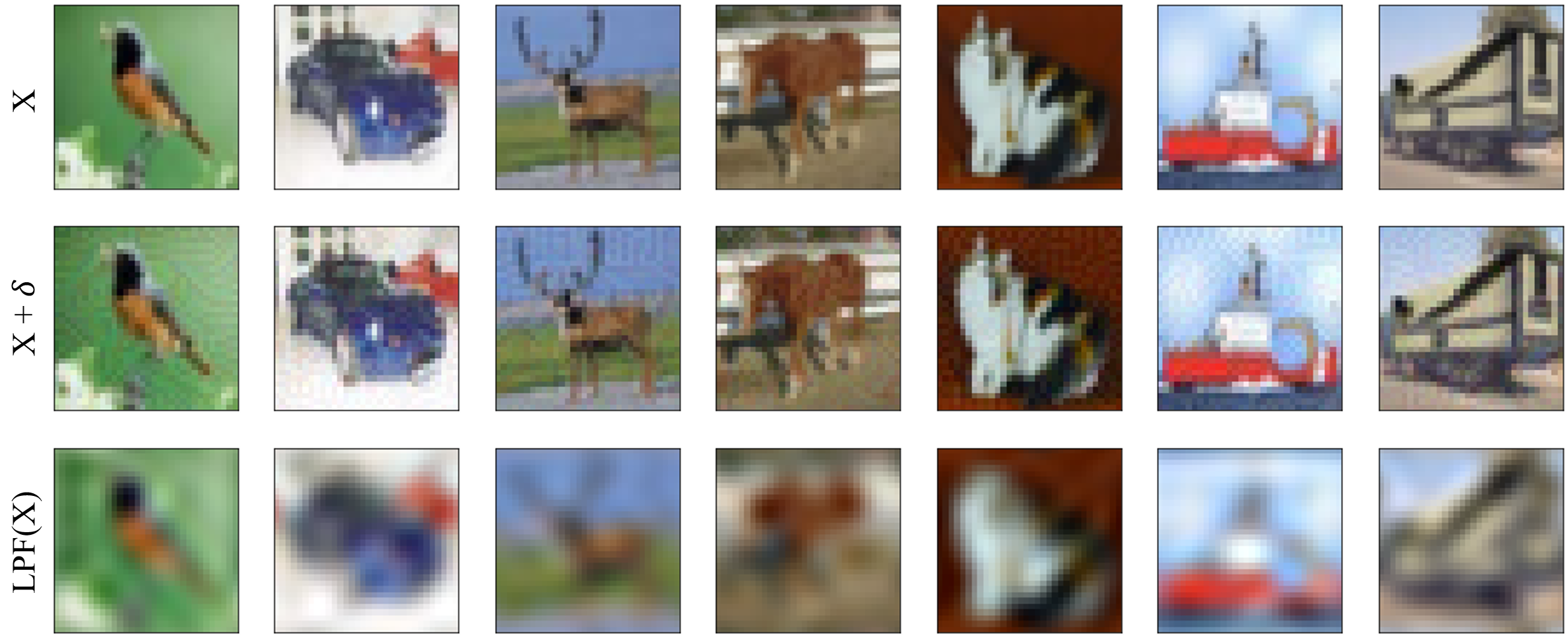}
\end{center}
\caption{Example CIFAR-10 natural images X (top), the perturbed images X+$\delta$ (middle), and LPF-processed images at a bandwidth of 8 (bottom).}
\label{lpf_x_prime}
\end{figure}

The above results highlight two key observations: \textbf{1)} AT prompts the model to focus primarily on extremely low frequencies to learn a more shape-biased representation for robustness. \textbf{2)} However, merely concentrating on the low-frequency region does not yield adversarial robustness, which explains why some data augmentations applied to the low-frequency region fail to increase robustness~\cite{yin2019fourier}.

\subsection{Frequency Distribution of Perturbations}
\label{frequencydistribution}

\paragraph{Frequency Distribution of Perturbations}
Frequency analysis provides a new lens on the understanding of network behaviors. %A deep understanding of the perturbation's frequency distribution can provide new insights towards the design of effective defensive methods. 
Indeed, some defenses are motivated specifically by the hypothesis that adversarial perturbations lie primarily in the high-frequency region. However, other researches~\cite{bernhard2021impact, maiya2021frequency} have refuted this hypothesis and claimed the adversarial frequency distribution is dataset-driven.

\begin{figure}[t]
\begin{center}
%\framebox[4.0in]{$\;$}
\includegraphics[width=0.98\linewidth]{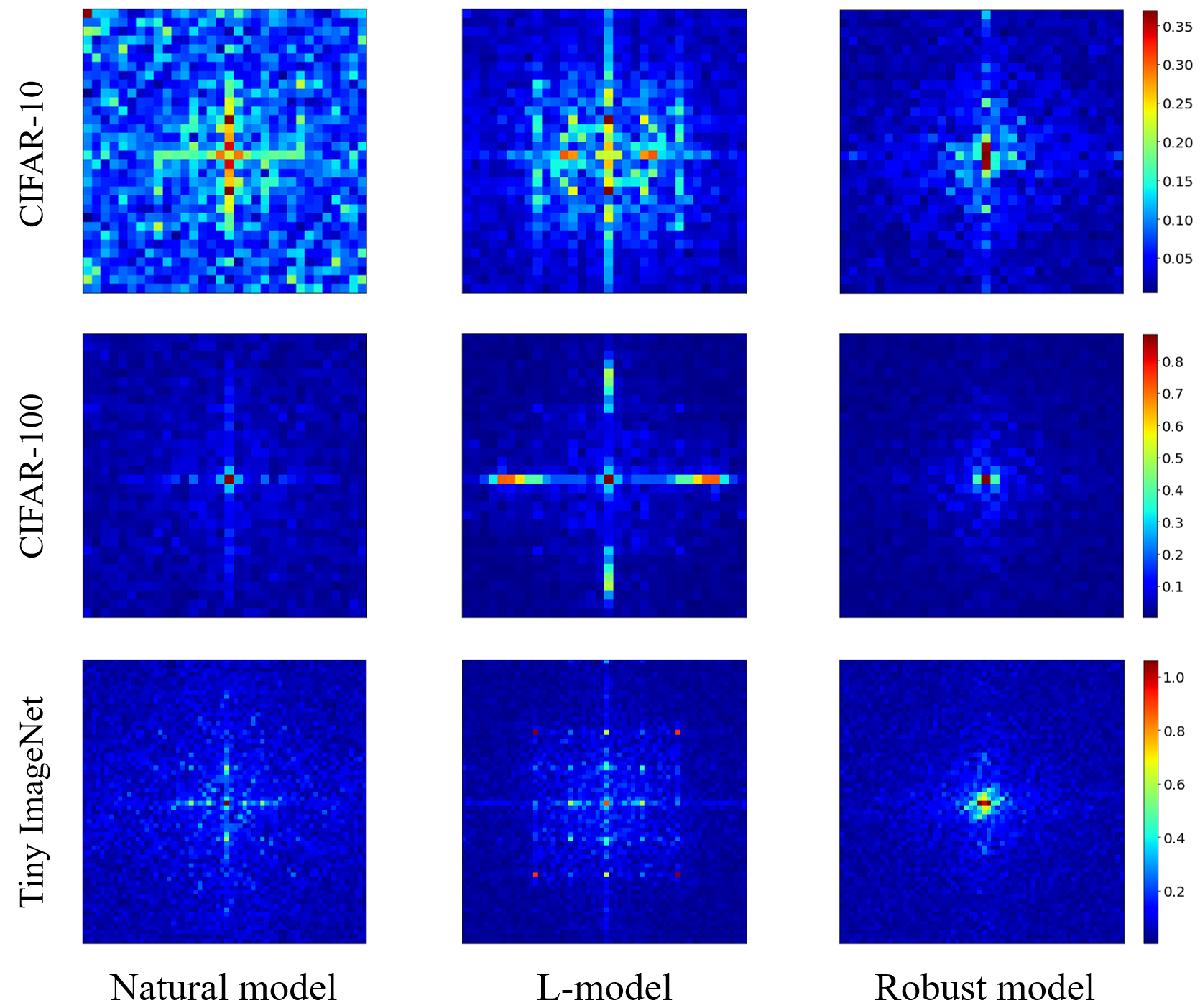}
\end{center}
\caption{Visualization of (2k-averaged) perturbations in the frequency domain,  with zero-frequency shifted to the center.}
\label{delta_fft}
\end{figure}

To this end, we use the PGD attack with the maximum perturbation $\epsilon = 8/255$ as a representative of the white-box attack and analyse the frequency distribution of the perturbations. Since the generation of perturbations is determined by two main components, namely, the target model and the input images, we hypothesize that \textit{the frequency distribution of the perturbations is related not only to the dataset but also to the model's frequency bias}. To verify this,
we take 2000 test examples to compute the average Fourier spectra of the PGD-20 attack perturbation for the natural, L- and robust models. The results are shown in Figure~\ref{delta_fft}.

\begin{figure}[th]
\begin{center}
%\framebox[4.0in]{$\;$}
\includegraphics[width=1.0\linewidth]{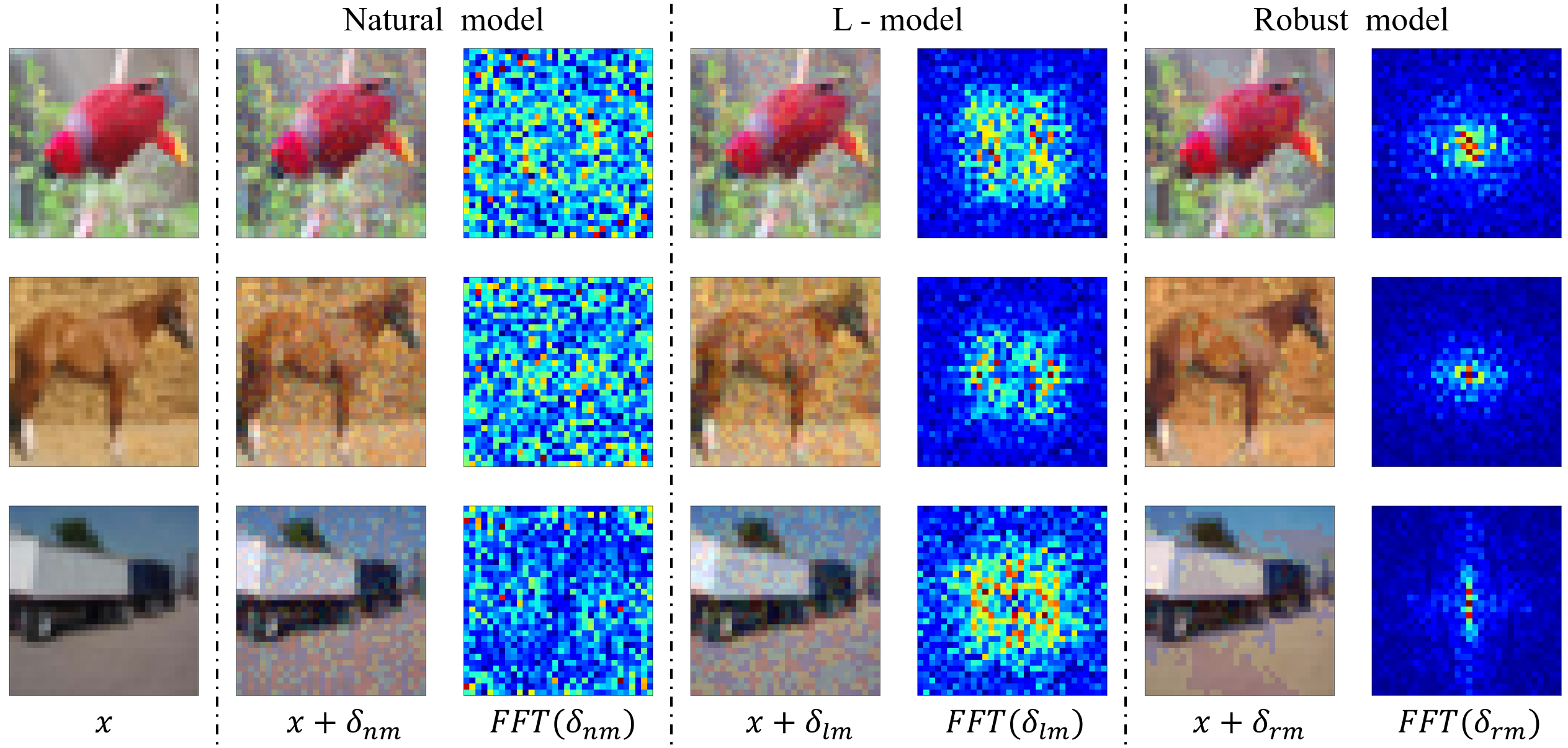}
\end{center}
\caption{Visualization of clean (leftmost) and attacked CIFAR-10 images (2-column pairs). $\delta_{nm}, \delta_{lm},$ and $\delta_{rm}$ denote the perturbations of natural, L- and robust models, respectively.}
\label{c10_delta}
\end{figure}

Across the datasets, for the natural and L-models, the frequency distribution of the perturbations differ significantly. Robust models bias the perturbations towards the low-frequency regions, which is consistent with the observation that robust models extract more information from low-frequency parts. The frequency distribution of the L-model is more concentrated in the low-frequency region compared to the natural one. There are also clear square contours in the CIFAR-10 and Tiny ImageNet datasets, indicating that the attacks are mainly concentrated within the central square. %High-amplitude values are largely concentrated in the centre (low-frequency region) for robust models.
These observations prove that the perturbation's frequency distribution is related to both the dataset and the model’s frequency bias. Furthermore, we also provide the frequency distributions of adversarial perturbations of randomly selected images on CIFAR-10 in Figure~\ref{c10_delta} to further verify this finding. The perturbations of natural, L- and robust models differ significantly in the spatial and frequency domains. More results across the datasets are shown in Appendix~\ref{FFT_delta_AE}.

\paragraph{Aggressiveness of Perturbation}
\begin{figure}[t]
\begin{center}
%\framebox[4.0in]{$\;$}
\includegraphics[width=0.97\linewidth]{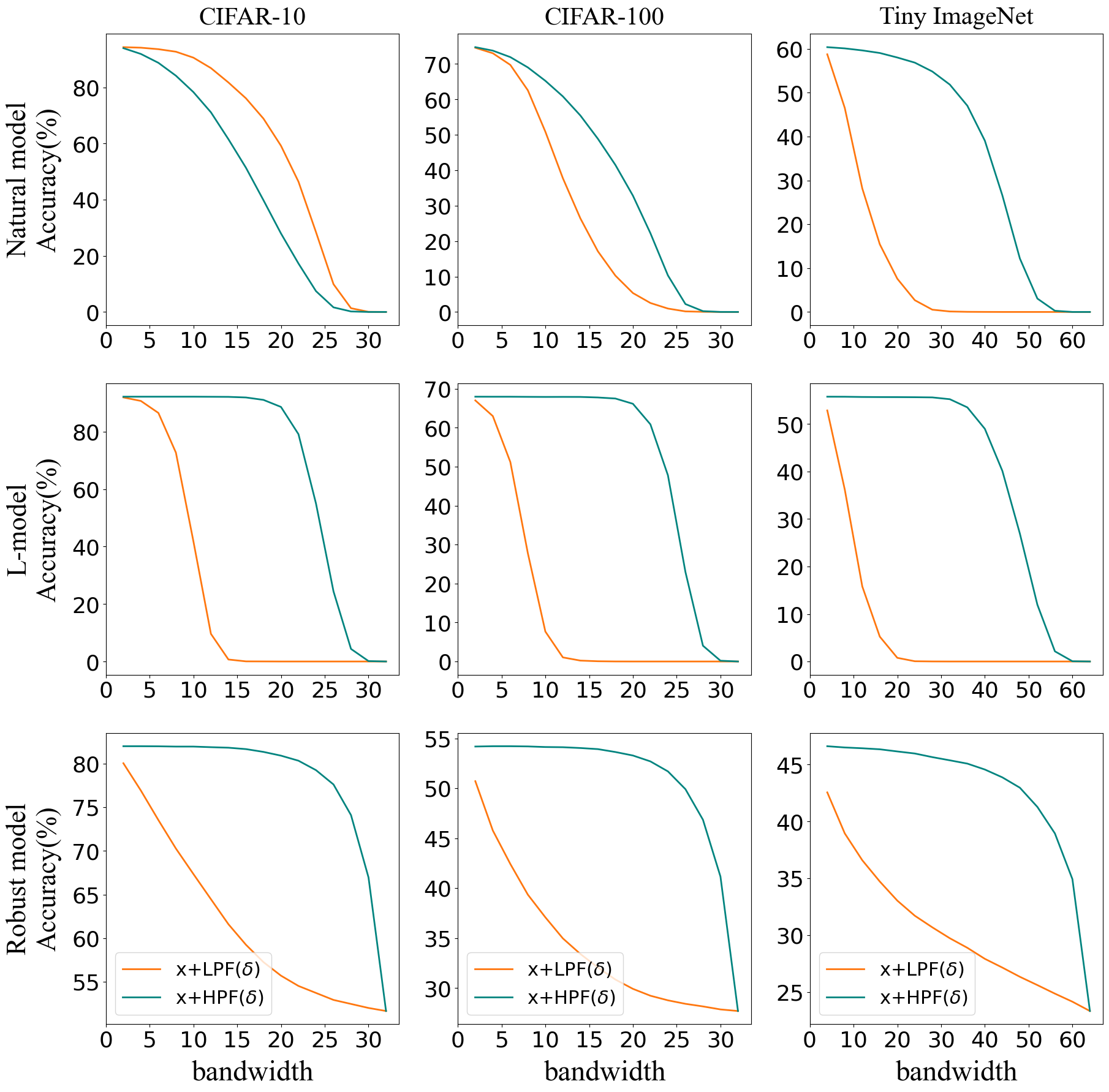}
\end{center}
\caption{Standard accuracy and robust accuracy against PGD-20 attacks for natural, L- and robust ResNet18 models across different datasets. As the bandwidth increases, more input or perturbation information is retained.}
\label{F_delta}
\end{figure}
\begin{figure}[th]
\begin{center}
%\framebox[4.0in]{$\;$}
\includegraphics[width=1.0\linewidth]{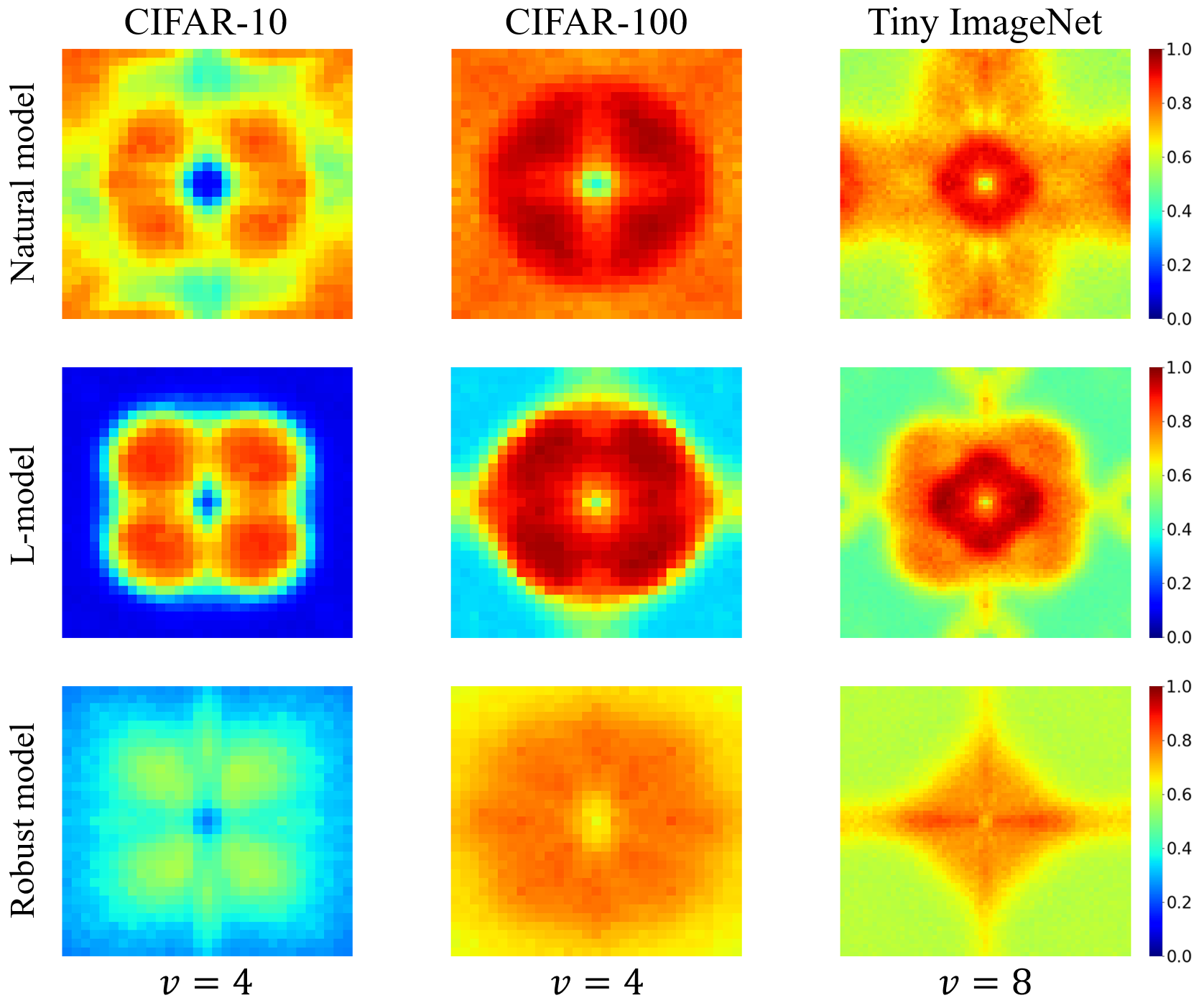}
\end{center}
\caption{Error rate of models on the images perturbed with the spectral perturbations (zero-frequency-centered). Hotter (red) regions denote higher fragility to attack. We evaluate three models on four datasets, with $v$ being the norm of the perturbation.}
\label{fourierheatmap}
\end{figure}
To further explore the perturbation's aggressiveness of different frequency bands, we apply LPF and HPF with different bandwidths to the perturbations and then add them to the natural inputs to check the robust accuracy performance. The greater the drop in robust accuracy, the more aggressive the perturbation is in that spectral band. The results are shown in Figure~\ref{F_delta}.

%SVHN is a unique dataset wherein the information is primarily concentrated in the low-frequency region. The models obtained on it by either training method rely mainly on low-frequency information for predictions. Perturbations also rely on low-frequency information to maintain their aggressiveness, while high-frequency perturbations barely degrade robust accuracy. 
For the natural models of CIFAR and Tiny ImageNet datasets, whether the perturbations are processed by LPF or HPF, the robust accuracy decreases as the bandwidth increases until it reaches almost zero. It indicates that perturbations maintain their aggressiveness in both low- and high-frequency parts, which corresponds to the fact that natural models use both low- and high-frequency information for predictions. In CIFAR-10, the perturbation after HPF (green curve) leads to more accuracy degradation compared to LPF (orange curve) at the same bandwidth, which means the high-frequency perturbation is more aggressive. The opposite is true in the CIFAR-100 and Tiny ImageNet. In particular, in Tiny ImageNet, the aggressiveness of the perturbation is mainly concentrated at low frequencies. For natural models utilizing low- and high-frequency information, why is the distribution of perturbation aggressiveness so different across different datasets? \textit{We propose a sensitivity hypothesis that white-box attacks can detect spectral bands where the model is sensitive and formulate the attack correspondingly.}

To prove this assumption, we investigate the sensitivity of models to frequency corruptions via the Fourier heat maps~\cite{yin2019fourier} shown in Figure~\ref{fourierheatmap}. The definition is described in Appendix~\ref{def_FHM}. A high error rate means that the model is vulnerable to the attacks with the corresponding frequency. The first row of Figure~\ref{fourierheatmap} shows the Fourier heat maps of natural models across different datasets. The CIFAR-10 and CIFAR-100 are sensitive to both low- and high-frequency perturbations, which is consistent with the phenomenon that the perturbations after LPF or HPF can significantly degrade the robust accuracy. Tiny ImageNet is much more vulnerable to the low-frequency perturbations, which corresponds to the fact that perturbation aggressiveness is mainly concentrated at low frequencies. These phenomena validate our assumption: \textit{white-box attacks can adapt to a model-sensitive band of the spectrum and attack it.}

For the L-models shown in Figure~\ref{F_delta}, which extract information from the low-frequency region, the perturbation relies heavily on low-frequency information to ensure the success of attacks. As for the robust models, in the CIFAR dataset, they relay mainly on low-frequency information for predictions, and the perturbation similarly relies on low-frequency parts to degrade the robust accuracy. For the Tiny ImageNet dataset, the model can improve standard accuracy by a small margin with the help of high-frequency information. There is a small decrease in robust accuracy in both the high bandwidth region for LPF and the low bandwidth region for HPF of the perturbation, indicating that the perturbation is somewhat aggressive at high frequencies. The aggressive frequency distribution of the perturbations basically corresponds to the model's attention to the frequency domain. The L-models and robust models are sensitive to the low-frequency perturbations shown in the second and the third rows of Figure~\ref{fourierheatmap}, which again concurs with our sensitivity hypothesis. In short, the white-box attacks can strike the frequency regions where the model's defenses are weak. Indeed, this is a first-ever spectral perspective to explain why white-box attacks are generally hard to defend. 

% CW attack has the similar performance
Based on the above experiments, we claim that the white-box attacks are primarily distributed in the frequency domain where the model attends to, and can adjust their aggressive frequency distribution according to the model's sensitivity to frequency corruptions. Comparing the Fourier heat maps of the natural and robust models on multiple datasets, it can be concluded that the robust model is less sensitive to spectral perturbations in most frequency regions. This suggests methods that reduce the model's sensitivity to frequency corruptions should help improve adversarial robustness. 

\section{Spectral Alignment Regularization}
Although AT improves robustness, there is still a large gap between robust and standard accuracies. Kannan \textit{et al.}~\cite{kannan2018adversarial} propose an adversarial logit pairing that forces the logits of a paired natural and adversarial examples to be similar.
%~\cite{mao2019metric} leverages metric learning to pull all the images of one class, both natural and adversarial, closer while pushing the images of other classes far apart.
%\textcolor{black}{In}~\cite{zhang2019theoretically}, \textcolor{black}{classification-calibrated loss is utilized} to minimize the difference between the prediction of natural and adversarial inputs.
\textcolor{black}{Bernhard \textit{et al.}}~\cite{bernhard2021impact} apply LPF and HPF to the inputs, and then minimize the output difference between natural and filtered inputs.
 Tack \textit{et al.}~\cite{tack2022consistency} improve the robustness by forcing the predictive distributions after attacking from two different augmentations of the same input to be similar.
% Based on the conclusion in Section~\ref{frequencydistribution} that the white-box perturbation can adapt its aggressive spectral bands to the target model, 
% \huang{we propose that the model will gain robustness if it could extract similar robust features regardless of frequency differences when faced with natural and adversarial input. Intuitively, a natural idea is to drive the outputs of natural and adversarial inputs to be as similar as possible in the frequency domain. By updating the weights through the back-propagation mechanism, this constraint makes the model extract similar frequency features from the adversarial inputs as the natural inputs. Then the robust accuracy will gradually approach the standard accuracy and thus be improved.}

\begin{figure}[h]
\begin{center}
%\framebox[4.0in]{$\;$}
\includegraphics[width=1.0\linewidth]{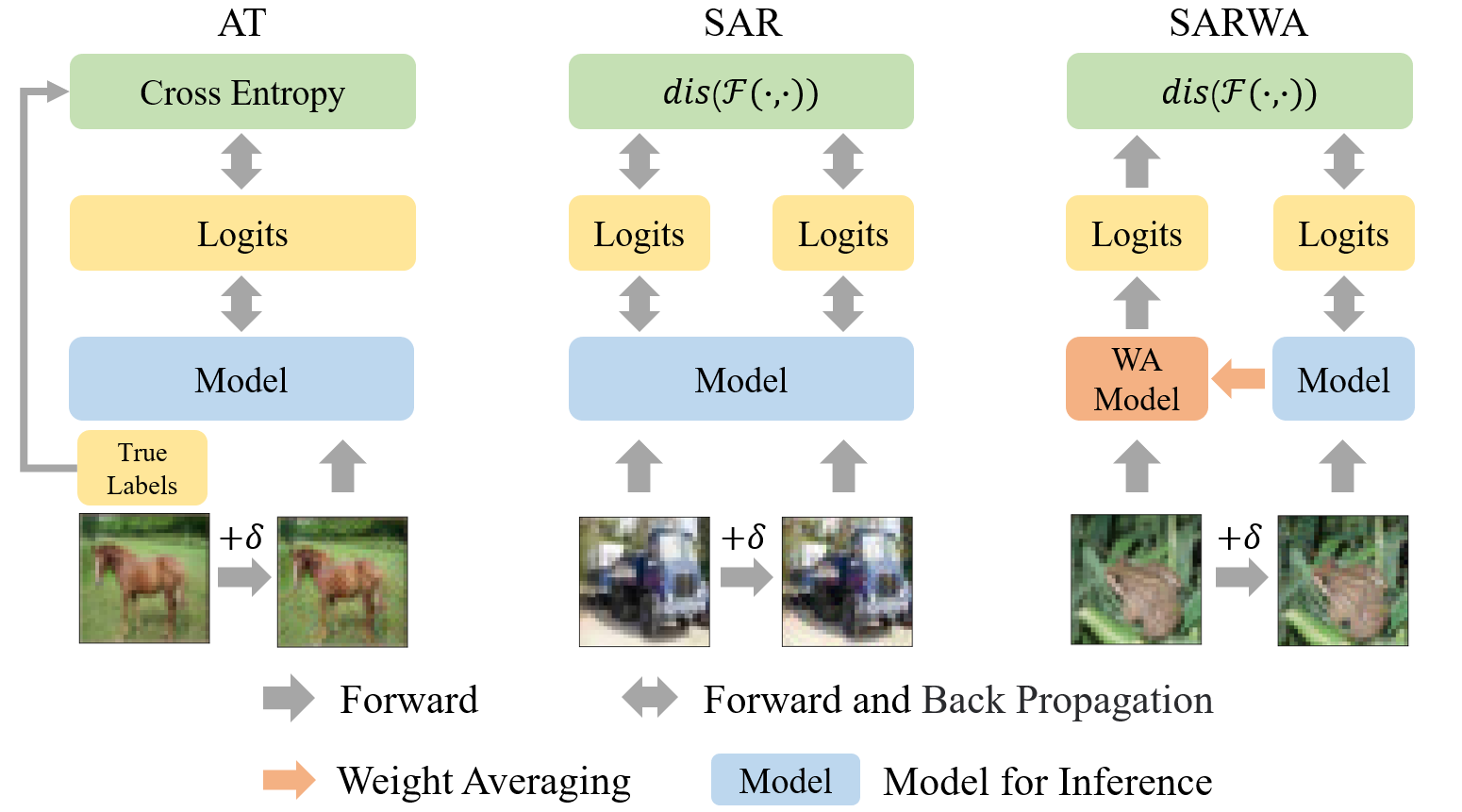}
\end{center}
\vspace{-4mm}
\caption{An overview of the standard AT, SAR, and SARWA. $\delta$ denotes the perturbation, $\mathcal{F}$ denotes the FFT, $\emph{dis}$ denotes the distance function.}
\label{flowchart}
\end{figure}

For the first time in literature, we have demonstrated in Section~\ref{frequencydistribution} that the frequency distribution of a perturbation is related to both the dataset and model, which is not a simple low- or high-frequency phenomenon, and the white-box attack can adapt its aggressive frequency distribution to the target model. To increase adversarial robustness, an intuitive idea is to drive the model to limit or tolerate this spectral difference between the outputs subject to a natural input and its adversarial counterpart. By updating the weights through the back-propagation mechanism, this constraint makes the model extract similar spectral features from the adversarial inputs as the natural inputs. Then the robust accuracy will gradually approach the standard accuracy and thus be improved. To achieve this goal, we devise a simple yet effective spectral alignment regularization (SAR) to align the difference of the outputs between the natural and adversarial inputs in the frequency domain, as shown in Figure~\ref{flowchart}. The optimization goal of the proposed AT with SAR is:
\begin{equation}
\label{FR}
\mathcal{L}_{AT} = \mathcal{L}_{CE} + \lambda \cdot \frac{1}{n} \sum_{i=1}^{n}\emph{Dis}\left(\mathcal{F}(f_{1}(x_{i})), \mathcal{F}(f_{2}(x_{i}+\delta))\right)
%\mathcal{C}_{2D} = \{(u, v)\; |\; (u, v) = f_{cal}(x, y, z), \; (x, y, z) \in %\mathcal{P}_{F} \}, 
\end{equation}
where $\lambda$ (defaulted to $0.1$) denotes the SAR coefficient, $f_{1}, f_{2}$ are the DNNs (same model for the SAR) and $f_{2}$ is used for prediction, $\emph{Dis}$ denotes the distance function ($\mathcal{L}_{1}$ norm is used), $\mathcal{L}_{CE}$ is the Cross-Entropy loss, and $\mathcal{F}$ denotes FFT. The distance function is applied to the real and imaginary parts of the complex numbers after FFT and then summed. SAR consists of two branches, one dealing with natural inputs and the other with adversarial inputs. Because the standard accuracy is higher than the robust accuracy, it may reduce the standard accuracy while increasing the robustness. To control the degradation of standard accuracy while maintaining robustness, we need to find the proper model to handle natural inputs. 

Weight averaging (WA)~\cite{izmailov2018averaging} (see Appendix~\ref{WA_eqn}) averages the weight values over epochs along the training trajectory, and is proved to be an effective means to improve model generalization. In AT, it could be combined with other methods~\cite{gowal2020uncovering, chen2020robust} to mitigate the robust overfitting problem~\cite{rice2020overfitting}. These works use WA to generate the final model for evaluation. During AT, the WA model maintains a similar standard and robust accuracy to the current training model, and its weights are fixed. If we utilize the WA model to deal with natural inputs, the branch of SARWA processing natural inputs (cf. WA branch in Figure~\ref{flowchart}) will not update the weights to force the standard accuracy toward the robust accuracy. Therefore, instead of using the WA model for the final evaluation, we utilize the WA model to deal with natural inputs. In simple terms, we replace the $f_{1}$ model in Eqn.~\ref{FR} with the WA model generated during AT.

%We can get two models in one training process, a standard model and a WA model, denoted as FR-WA1 and FR-WA2. 

\section{Experiments}
\paragraph{Datasets}
Without loss of generality, we select three common image datasets: %SVHN~\cite{netzer2011reading},
CIFAR-10~\cite{krizhevsky2009learning}, CIFAR-100~\cite{krizhevsky2009learning} and Tiny ImageNet~\cite{le2015tiny}. For a fair comparison, all experiments adopt the same data augmentation method: 4-pixel padding with 32 × 32 random crops (except for Tiny ImageNet) and random horizontal flip. The value of the pixels in all natural images are normalized to be in the interval $[0,1]$. The resolution of each image in the CIFAR datasets is $32\times 32\times 3$, corresponding to the length, width, and channel, respectively. Tiny ImageNet image resolution is $64\times 64 \times 3$. The SAR coefficient is set to 0.1 (0.15)\footnote{The number in brackets is the coefficient for SARWA} for CIFAR datasets, and 0.05 for Tiny ImageNet. The training set was randomly divided into the training set and the validation set according to the ratio of 9:1. We select the model with the highest robustness against PGD-20 for further evaluation on the validation set against other popular attacks.

\paragraph{Experimental Settings}
We take ResNet18 as a default model and adopt a SGD optimizer with a momentum of 0.9 and a weight decay of $5\times10^{-4}$. The model is trained with PGD-10 for 100 epochs with a batch size of 128 on one 3090 GPU. The initial learning rate is 0.1, which decays to one-tenth at 75th and 90th epochs, respectively. The robust accuracy of the PGD-20 with a random-start is taken as the main basis for robustness analysis. The attack step size is $\alpha$ = 2/255 and maximum $l_\infty$ norm-bounded perturbation $\epsilon$ = 8/255. SAR and WA are used since the first epoch where the learning rate drops, and continues until the end with a cycle length 1.

\paragraph{Evaluated Attacks}
The model with the highest robust accuracy against PGD-20 is selected for further evaluation. To avoid a false sense of security caused by the obfuscated gradients, we evaluate the robust accuracy against several popular white-box attack methods, including PGD~\cite{madry2017towards}, C\&W~\cite{carlini2017towards}, and AutoAttack~\cite{croce2020reliable} (denoted as AA, consists of APGD-CE, APGD-DLR, FAB, and Square). Following the default setting of AT, the attack step size is 2/255, and the maximum $l_\infty$ norm-bounded perturbation is 8/255. 
% The standard and robust accuracies are used as the evaluation metrics.

\subsection{Experimental Results and Analyses}
\paragraph{Superior Performance across Datasets} As shown in Table~\ref{robust_accuracy}, we incorporate SAR and SARWA into AT to improve the robust accuracy against various attacks on multiple datasets. In particular, for AT, the SARWA version improves 3.14\% and 1.85\% of robust accuracy on average against the PGD-20 and AA attacks, respectively, with a much smaller degradation (0.28\%) in standard accuracy. These results indicate that our methods are versatile across various datasets. 

\paragraph{Benchmark with Other Defenses} Table~\ref{wrn} further compares the impact of SAR and SARWA with famous defenses (details of the defenses are reviewed in Appendix~\ref{popular}) on the CIFAR-10 dataset. Wide ResNet-34-10~\cite{zagoruyko2016wide} is the popular model for comparison. The results show that SAR and SARWA substantially improve robust accuracy compared to AT, and outperform other defenses. In particular, SARWA improves robustness while up-keeping standard accuracy, which is non-trivial, since there is a trade-off between the standard and robust accuracies. 

\begin{table}[t]
\caption{Top-1 accuracy(\%) against diverse attacks with maximum $l_\infty$ norm-bounded perturbation $\epsilon$ = 8/255 of ResNet18 models. Bold \textbf{\textcolor{red}{red}} and \textbf{\textcolor{blue}{blue}} numbers indicate the top and runner-up on different datasets.}
\label{robust_accuracy}
\begin{center}
\resizebox{\linewidth}{!}{
\begin{tabular}{c|c|ccccc}
\hline
%\multicolumn{1}{c}{Dataset}  & \multicolumn{1}{c}{PGD-20}
Dataset & Method & Clean & PGD-20 & PGD-50 & C\&W & AA 
\\ \hline 
% \multirow{3}{*}{SVHN} 
% & PGD-AT    & \textbf{\textcolor{red}{90.88}} & 53.28 & 52.26 & 50.62 & 47.57\\
% & AT+SAR     & 90.43 & \textbf{\textcolor{blue}{56.87}} & \textbf{\textcolor{blue}{56.29}} & \textbf{\textcolor{red}{51.89}} & \textbf{\textcolor{red}{49.46}}\\
% & AT+SARWA  & \textbf{\textcolor{blue}{90.49}} & \textbf{\textcolor{red}{56.95}} & \textbf{\textcolor{red}{56.35}} & \textbf{\textcolor{blue}{51.76}} & \textbf{\textcolor{blue}{49.36}}\\
% \hline

\multirow{3}{*}{CIFAR-10}  
& PGD-AT    & \textbf{\textcolor{red}{81.98}} & 51.69 & 51.46 & 50.44 & 48.19\\
& AT+SAR     & 80.04 & \textbf{\textcolor{red}{55.39}} & \textbf{\textcolor{red}{55.13}} & \textbf{\textcolor{blue}{51.61}} & \textbf{\textcolor{blue}{50.02}}\\
& AT+SARWA  & \textbf{\textcolor{blue}{81.74}} & \textbf{\textcolor{blue}{55.12}} & \textbf{\textcolor{blue}{54.88}} & \textbf{\textcolor{red}{52.21}} & \textbf{\textcolor{red}{50.16}}\\

\hline 
\multirow{3}{*}{CIFAR-100} 
& PGD-AT    & \textbf{\textcolor{red}{54.18}} & 27.81 & 27.49 & 25.82 & 23.77\\
& AT+SAR     & 49.23 & \textbf{\textcolor{blue}{31.27}} & \textbf{\textcolor{blue}{31.20}} & \textbf{\textcolor{blue}{27.60}} & \textbf{\textcolor{red}{26.09}}\\
& AT+SARWA & \textbf{\textcolor{blue}{53.66}} & \textbf{\textcolor{red}{31.49}} & \textbf{\textcolor{red}{31.36}} & \textbf{\textcolor{red}{28.24}}  & \textbf{\textcolor{blue}{26.06}}\\

\hline
\multirow{3}{*}{Tiny ImageNet} 
& PGD-AT    & \textbf{\textcolor{red}{46.64}} & 23.33 & 23.18 & 20.44 & 18.34\\
& AT+SAR     & 42.92 & \textbf{\textcolor{blue}{25.32}} & \textbf{\textcolor{blue}{25.25}} & \textbf{\textcolor{blue}{21.58}} & \textbf{\textcolor{red}{19.71}} \\
& AT+SARWA & \textbf{\textcolor{blue}{46.55}} & \textbf{\textcolor{red}{25.64}} & \textbf{\textcolor{red}{25.47}} & \textbf{\textcolor{red}{21.90}} & \textbf{\textcolor{blue}{19.64}}\\
\hline 

\end{tabular}}
\end{center}
\end{table}

\begin{table}[t]
\caption{Top-1 accuracy(\%) of the WideResNet-34-10 on the CIFAR-10. Bold \textbf{\textcolor{red}{red}} and \textbf{\textcolor{blue}{blue}} numbers indicate the top and runner-up.}
\label{wrn}
\begin{center}
\resizebox{\linewidth}{!}{
\begin{tabular}{c|ccccc}
\hline
%\multicolumn{1}{c}{Dataset}  & \multicolumn{1}{c}{PGD-20}
Method  & Clean & PGD-20 & PGD-50 & C\&W & AA 
\\ \hline 
PGD-AT~\cite{rice2020overfitting} & 84.62 & 55.01 & 54.88 & 53.32 & 51.42  \\

%ALP~\cite{kannan2018adversarial} & 83.52 & 56.9 & 56.71 & 54.31 & 52.78\\ 

TRADES~\cite{zhang2019theoretically} & \textbf{\textcolor{blue}{84.92}} & 56.33 & 56.13 & 54.20 & 53.08  \\ 

MART~\cite{wang2019improving} & 84.17 & 58.56 & 58.06 & 54.58 & 51.10  \\ 

AWP~\cite{wu2020adversarial} & \textbf{\textcolor{red}{85.57}} & 58.14 & 57.92 & \textbf{\textcolor{blue}{55.96}} & 54.04  \\
%AT-SWA & \textbf{86.17} & 55.18 &  & 54.57 & 52.25 \\

% \\ \hline 
% PGD-AT~\cite{rice2020overfitting} & 84.62 & 55.01 & 54.88 & 53.32 & 51.42  \\

% %ALP~\cite{kannan2018adversarial} & 83.52 & 56.9 & 56.71 & 54.31 & 52.78\\ 

% TRADES~\cite{zhang2019theoretically} & 84.92 & 56.33 & 56.13 & 54.20 & 53.08  \\ 

% MART~\cite{wang2019improving} & 84.17 & 58.56 & 58.06 & 54.58 & 51.10  \\ 

% AWP~\cite{wu2020adversarial} & \textbf{85.57} & 58.14 & 57.92 & 55.96 & 54.04  \\
% %AT-SWA & \textbf{86.17} & 55.18 &  & 54.57 & 52.25 \\

\hline

SAR (ours) & 82.66 & \textbf{\textcolor{red}{59.38}} & \textbf{\textcolor{red}{59.15}} & 55.72 & \textbf{\textcolor{blue}{54.33}} \\
SARWA (ours) & 84.87 & \textbf{\textcolor{blue}{58.79}} & \textbf{\textcolor{blue}{58.50}} & \textbf{\textcolor{red}{56.23}} & \textbf{\textcolor{red}{54.35}} \\

\hline 

\end{tabular}}
\end{center}
\end{table}

\paragraph{Plug-and-play blocks}
Since SAR and SARWA are plug-and-play blocks, we also apply them to popular defense techniques to further improve the robust accuracy. The thorough performances across multiple defense methods on ResNet18 are shown in Table~\ref{cifar10_more}. More experiments across the datasets in Appendix~\ref{more_experiments} indicate the effectiveness of the proposed SAR and SARWA methods. Experimental results demonstrate that SAR and SARWA can be plugged into these popular methods to further improve robustness. Besides, SARWA can maintain a similar standard accuracy as the original defense while improving robust accuracy.

\begin{table}[ht]
\caption{Top-1 accuracy(\%) of the ResNet18 model on the CIFAR-10. Bold \textbf{\textcolor{red}{red}} and \textbf{\textcolor{blue}{blue}} numbers indicate the top and runner-up.}
\label{cifar10_more}
\begin{center}
\resizebox{\linewidth}{!}{
\begin{tabular}{c|ccccc}
\hline
%\multicolumn{1}{c}{Dataset}  & \multicolumn{1}{c}{PGD-20}
Method  & Clean & PGD-20 & PGD-50 & C\&W & AA 
\\ \hline 
AT & \textbf{\textcolor{red}{81.98}} & 51.69 & 51.46 & 50.44 & 48.19\\

SAR & 80.04 & \textbf{\textcolor{red}{55.39}} & \textbf{\textcolor{red}{55.13}} & \textbf{\textcolor{blue}{51.61}} & \textbf{\textcolor{blue}{50.02}}\\

SARWA & \textbf{\textcolor{blue}{81.74}} & \textbf{\textcolor{blue}{55.12}} & \textbf{\textcolor{blue}{54.88}} & \textbf{\textcolor{red}{52.21}} & \textbf{\textcolor{red}{50.16}}\\

\hline 
TRADES & \textbf{\textcolor{red}{81.83}} & 53.41 & 53.23 & 50.92 & 49.84 \\

TRADES + SAR & 80.11 & \textbf{\textcolor{red}{54.45}} & \textbf{\textcolor{red}{54.24}} & \textbf{\textcolor{red}{51.51}} & \textbf{\textcolor{red}{50.52}} \\

TRADES + SARWA & \textbf{\textcolor{blue}{81.76}} & \textbf{\textcolor{blue}{54.14}} & \textbf{\textcolor{blue}{54.06}} & \textbf{\textcolor{blue}{51.31}} & \textbf{\textcolor{blue}{50.38}} \\

\hline 
MART & \textbf{\textcolor{red}{81.01}} & 54.58 & 54.47 & 50.01 & 48.10 \\ %CW 54.87

MART + SAR & 79.03 & \textbf{\textcolor{red}{55.17}} & \textbf{\textcolor{red}{54.90}}& \textbf{\textcolor{red}{50.98}} & \textbf{\textcolor{red}{49.22}} \\

MART + SARWA & \textbf{\textcolor{blue}{80.80}} & \textbf{\textcolor{blue}{55.01}} & \textbf{\textcolor{blue}{54.78}} & \textbf{\textcolor{blue}{50.12}} & \textbf{\textcolor{blue}{48.78}} \\

\hline 
AWP & \textbf{\textcolor{red}{81.06}} & 55.36 & 55.27 & 51.98 & 50.37 \\

AWP + SAR & 79.03 & \textbf{\textcolor{red}{57.07}} & \textbf{\textcolor{red}{57.01}} & \textbf{\textcolor{red}{52.16}} & \textbf{\textcolor{red}{50.80}} \\

AWP + SARWA & \textbf{\textcolor{blue}{80.87}} & \textbf{\textcolor{blue}{56.81}} & \textbf{\textcolor{blue}{56.82}} & \textbf{\textcolor{blue}{52.14}} & \textbf{\textcolor{blue}{50.61}} \\

\hline

\end{tabular}}
\end{center}
\end{table}

\paragraph{Large models}
\label{large_models}
To prove the effectiveness of the SAR and SARWA blocks on large models, we trained the ResNet152 and Wide ResNet-34-15 on the CIFAR-10 dataset. As shown in Table~\ref{large_model}. SAR and SARWA achieve a higher adversarial robustness against multiple well-recognized attacks relative to the AT, while SARWA can maintain a similar standard accuracy on the CIFAR-10 dataset. The improvement is non-trivial, since some papers have claimed a trade-off between the standard and robust accuracies~\cite{tsipras2018robustness, zhang2019theoretically}, demonstrating the effectiveness and feasibility of the proposed SAR and SARWA methods.

\begin{table}[ht]
% \vspace{8mm}
\caption{Top-1 accuracy(\%) of various models on the CIFAR-10. \#number indicates the parameters. Bold \textbf{\textcolor{red}{red}} and \textbf{\textcolor{blue}{blue}} numbers indicate the top and runner-up.}
\label{large_model}
\begin{center}
\resizebox{\linewidth}{!}{
\begin{tabular}{c|c|ccccc}
\hline
%\multicolumn{1}{c}{Dataset}  & \multicolumn{1}{c}{PGD-20}
model & Method  & Clean & PGD-20 & PGD-50 & C\&W & AA 
\\ \hline \multirow{3}{*}{\makecell{ResNet152 \\ \# 58,156,618}}
 & AT    & \textbf{\textcolor{blue}{84.60}} & 54.62 & 54.44 & 52.86 & 50.87\\
 & SAR  & 81.11 & \textbf{\textcolor{red}{57.76}} & \textbf{\textcolor{red}{57.70}} & \textbf{\textcolor{blue}{54.74}} & \textbf{\textcolor{blue}{53.32}}\\
 & SARWA & \textbf{\textcolor{red}{84.62}} & \textbf{\textcolor{blue}{57.55}} & \textbf{\textcolor{blue}{57.50}} & \textbf{\textcolor{red}{55.63}} & \textbf{\textcolor{red}{54.10}}\\
 
\hline \multirow{3}{*}{\makecell{Wide ResNet-34-15 \\ \# 103,819,674}}
& AT & \textbf{\textcolor{red}{86.66}} & 55.79 & 55.49 & 54.58 & 52.76 \\

& SAR & 84.35 & \textbf{\textcolor{red}{58.33}} & \textbf{\textcolor{red}{58.03}} & \textbf{\textcolor{red}{55.75}} & \textbf{\textcolor{red}{54.12}} \\

& SARWA & \textbf{\textcolor{blue}{86.18}} & \textbf{\textcolor{blue}{57.97}} & \textbf{\textcolor{blue}{57.78}} & \textbf{\textcolor{blue}{55.25}} & \textbf{\textcolor{blue}{53.50}} \\
\hline 
\end{tabular}}
\end{center}
\end{table}

\subsection{Ablation Studies}
\begin{figure}[h]
\begin{center}
%\framebox[4.0in]{$\;$}
% figure 1
\subfigure[Standard Accuracy]{
\begin{minipage}[t]{0.45\linewidth}
\centering
\label{dis1}
\includegraphics[width=1.0\linewidth]{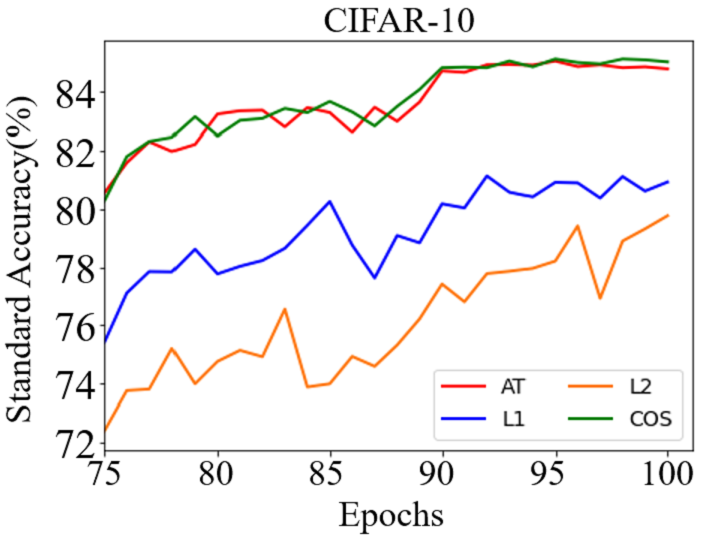}
\end{minipage}
}
\subfigure[Robust Accuracy]{
\begin{minipage}[t]{0.45\linewidth}
\centering
\label{dis2}
\includegraphics[width=1.0\linewidth]{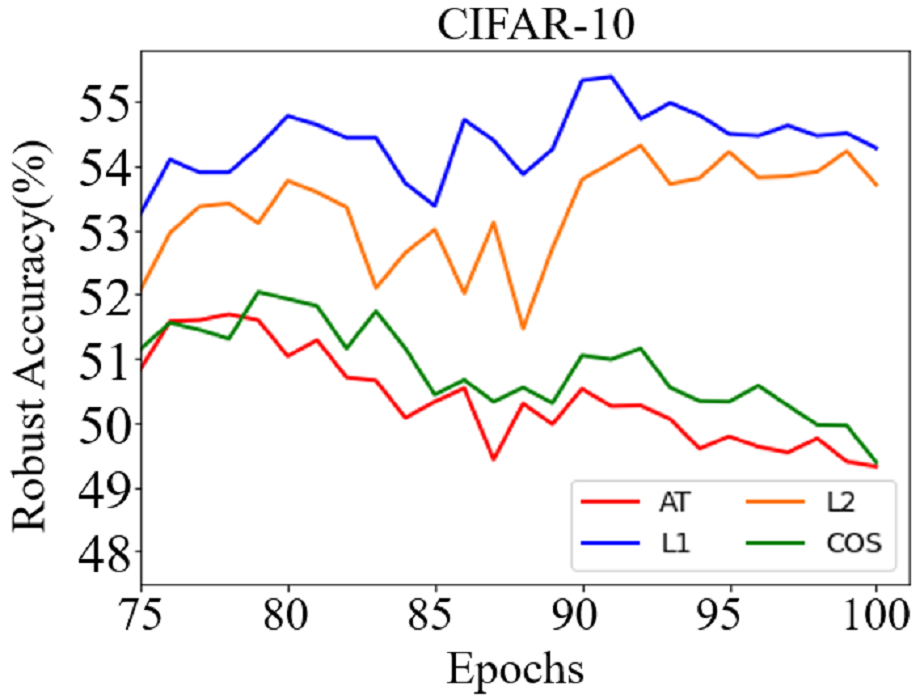}
\end{minipage}
}

\end{center}
% \vspace{-4mm}
\caption{Ablation of distance functions on CIFAR-10. (a) and (b) show the standard and robust accuracy of different distance functions, AT represents the standard adversarial training without SAR. }
\label{ablation}
\end{figure}

\paragraph{Distance Function in SAR}
For the distance function defined in Eqn.~\ref{FR}, there are three frequently used methods, including $\mathcal{L}_{1}$ norm, $\mathcal{L}_{2}$ norm, and cosine similarity. We measure their performance on CIFAR-10. For fair comparison, we use the same checkpoint from the 74th epoch and then apply the different distance functions, respectively.
As shown in Figures~\ref{dis1} and~\ref{dis2}, in terms of improving robustness, $\mathcal{L}_{1}$ norm is more effective at the expense of a slightly reduced standard accuracy.

\begin{figure}[h]
\label{hyper}
\begin{center}
%\framebox[4.0in]{$\;$}
% figure 1
\subfigure[SAR]{
\begin{minipage}[t]{0.46\linewidth}
\centering
\label{hyper_SAR}
\includegraphics[width=1.0\linewidth]{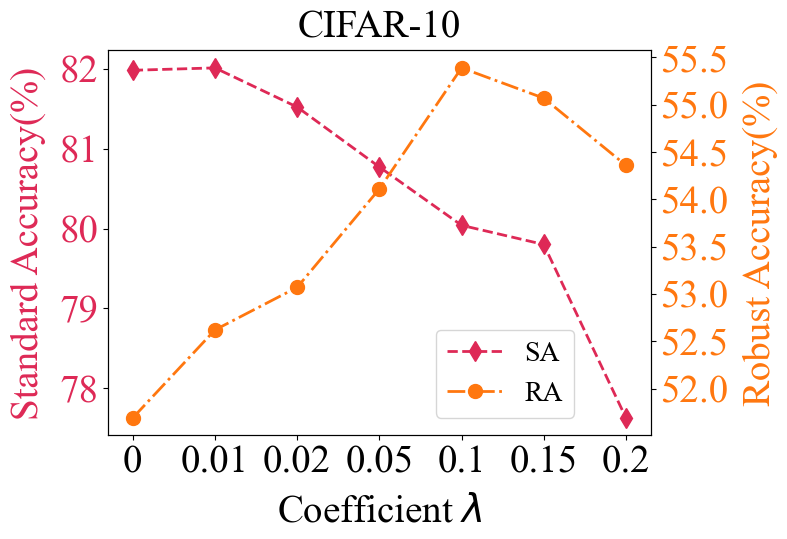}
\end{minipage}
}
\subfigure[SARWA]{
\begin{minipage}[t]{0.46\linewidth}
\centering
\label{hyper_SARWA}
\includegraphics[width=1.0\linewidth]{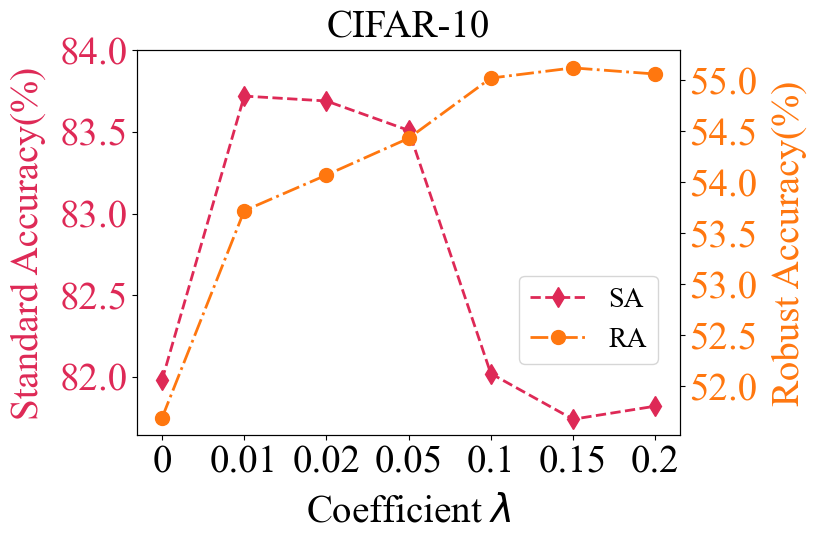}
\end{minipage}
}

\end{center}
% \vspace{-4mm}
\caption{Ablation of coefficient $\lambda$ of SAR and SARWA on CIFAR-10. $\lambda=0$ represents the standard AT without SAR or SARWA. }
\label{ablation2}
\end{figure}

\paragraph{Sensitivity to Regularization Coefficient}
We investigate the impact of the super parameter $\lambda$ in SAR defined in Eqn.~\ref{FR}, which modifies the strength of the spectral alignment regularization. The results for different $\lambda \in [0, 0.2]$ are shown in Figures~\ref{hyper_SAR} and~\ref{hyper_SARWA}. $0$ represents the standard AT. Results show a trade-off between the standard and robust accuracies when the $\lambda$ is within a proper limit $[0, 0.1]$ for SAR. When $\lambda$ is outside this range, SAR dominates the loss function, resulting in an overall decrease in both standard and robust accuracies. Compared to AT, SARWA can improve both standard and robust accuracy with a small $\lambda$, which is a noteworthy achievement. As our primary objective is to prioritize robustness, we select the $\lambda$ value (e.g., 0.15) that yields the highest robust accuracy for comparing against other defense methods.

%When $\lambda$ is outside this range, SAR dominates the loss function and diverts the optimization target from the classification task, resulting in an overall decrease in both standard and robust accuracies.

\section{Conclusion}
This work explores the appealing properties of adversarial perturbation and adversarial training (AT) from a spectral lens. We find that AT renders the model more focused on shape-biased representation in the low-frequency region to gain robustness. 
%We use the empirical results to show that the frequency distribution of perturbations depends on the dataset and the target model.
Using systematic experiments, we show for the first time that the white-box attack can adapt its aggressive frequency distribution to the target model's sensitivity to frequency corruptions, making it hard to defend.  
These novel insights advance our knowledge about the frequency mechanism of AT, and inspire us to devise a spectral alignment regularization (SAR) for aligning the spectral outputs with respect to natural and adversarial inputs. Experiments then verify that SAR substantially improves robust accuracy against frequency-varying perturbations, without extra data. 
% Our future work will focus on reducing the model's sensitivity.

\section*{Acknowledgments}
This work is supported in part by the General Research Fund (GRF) project 17209721, and in part by the Theme-based Research Scheme (TRS) project T45-701/22-R of the Research Grants Council (RGC), Hong Kong SAR.


{\small
\bibliographystyle{IEEEtran} 
\bibliography{tnnls}
}

\ifCLASSOPTIONcaptionsoff
  \newpage
\fi

\clearpage
\appendix
\section{Appendix}
\subsection{Visualization of the natural, perturbed and LPF-processed images from different datasets.}
\label{app1}
% \begin{figure}[ht]
% \begin{center}
% %\framebox[4.0in]{$\;$}
% \includegraphics[width=1.0\linewidth]{Fig/svhn_lpf_x_prime_.png}
% \end{center}
% \caption{Visualization of the SVHN images after LPF with a bandwidth of 8 (top), natural images X (middle), and the perturbed images X+$\delta$ (bottom).}
% \label{svhn}
% \end{figure}

\begin{figure}[ht]
\begin{center}
%\framebox[4.0in]{$\;$}
\includegraphics[width=1.0\linewidth]{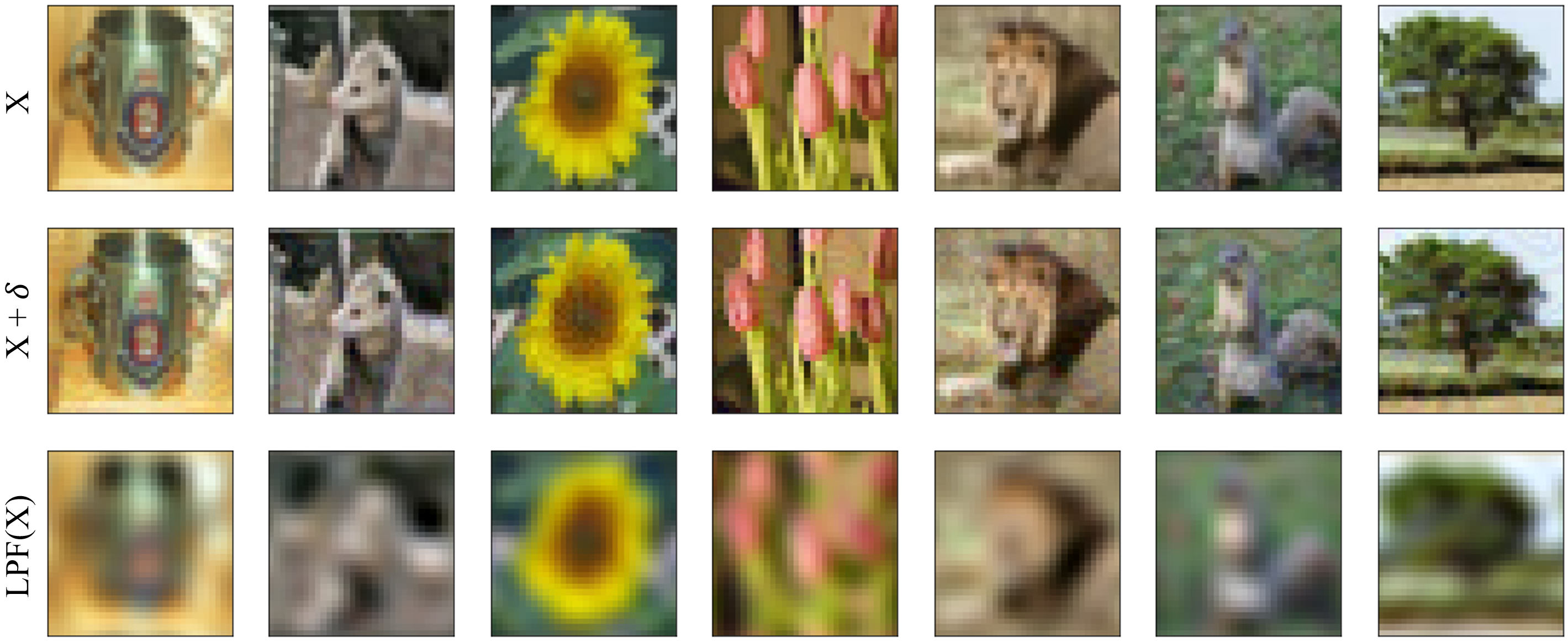}
\end{center}
\caption{Example CIFAR-100 natural images X (top), the perturbed images X+$\delta$ (middle), and LPF-processed images at a bandwidth of 8 (bottom).}
\label{c100}
\end{figure}

\begin{figure}[ht]
\begin{center}
%\framebox[4.0in]{$\;$}
\includegraphics[width=1.0\linewidth]{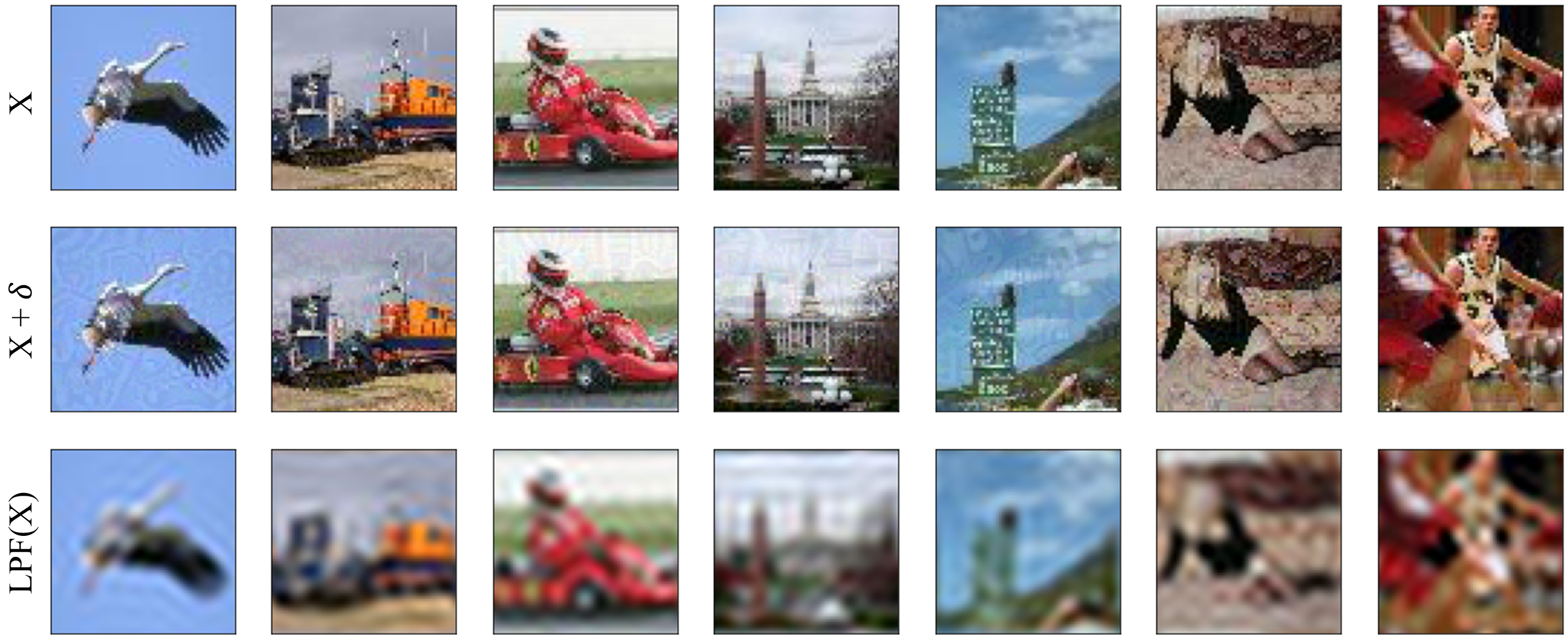}
\end{center}
\caption{Example Tiny ImageNet natural images X (top), the perturbed images X+$\delta$ (middle), and LPF-processed images at a bandwidth of 8 (bottom).}
\label{tin}
\end{figure}

\subsection{Fourier Heat Map}
\label{def_FHM}
Fourier heat map~\cite{yin2019fourier} provides a perturbation analysis method to investigate the sensitivity of models to the frequency corruptions. More precisely, let $\mathbf{U}_{i,j} \in \mathbb{R}^{d_{1} \times d_{2}}$ be a real-valued matrix (2D Fourier basis matrices) such that $\|\mathbf{U}_{i,j}\|_{2}=1$, and $FFT(\mathbf{U}_{i,j})$ only has up to two non-zero elements located at $(i,j)$ and its symmetric coordinate with respect to the center. FFT denotes Fast Fourier Transform.
Given a model, we can generate the perturbed image $\widetilde{X} = X + rv\mathbf{U}_{i,j}$ from the natural image $X$, where $r$ is chosen uniformly at random from $\{-1,1\}$, and $v$ is the norm magnitude of the perturbation. For multi-channel images, we perturb every channel independently. We can then calculate the error rate of the model under Fourier basis noises and visualize how the error rate changes as a function of the spectral indices. The visualization result is called a Fourier heat map. In this paper, we move the low-frequency region to the center of the image. A high error rate means the model is vulnerable to attacks with the corresponding frequency.

\subsection{Weight Averaging}
\label{WA_eqn}
Following the definition in~\cite{izmailov2018averaging}, the equation of WA is:
\begin{equation}
\mathcal{W}^{n}_{wa} = \frac{\mathcal{W}^{n-1}_{wa} \times k + \mathcal{W}^{n}}{k+1}
\end{equation}
where $k$ denotes the number of past checkpoints to be averaged, $n$ denotes the index of the epoch during the training, $\mathcal{W}^{n}_{wa}$ denotes the weights of the WA model at $n$-th epoch, $\mathcal{W}^{n}$ denotes the current model's weights.

\subsection{Adversarial Perturbations and Spectral Distribution}
\label{FFT_delta_AE}
\begin{figure}[ht]
\begin{center}
%\framebox[4.0in]{$\;$}
\includegraphics[width=0.95\linewidth]{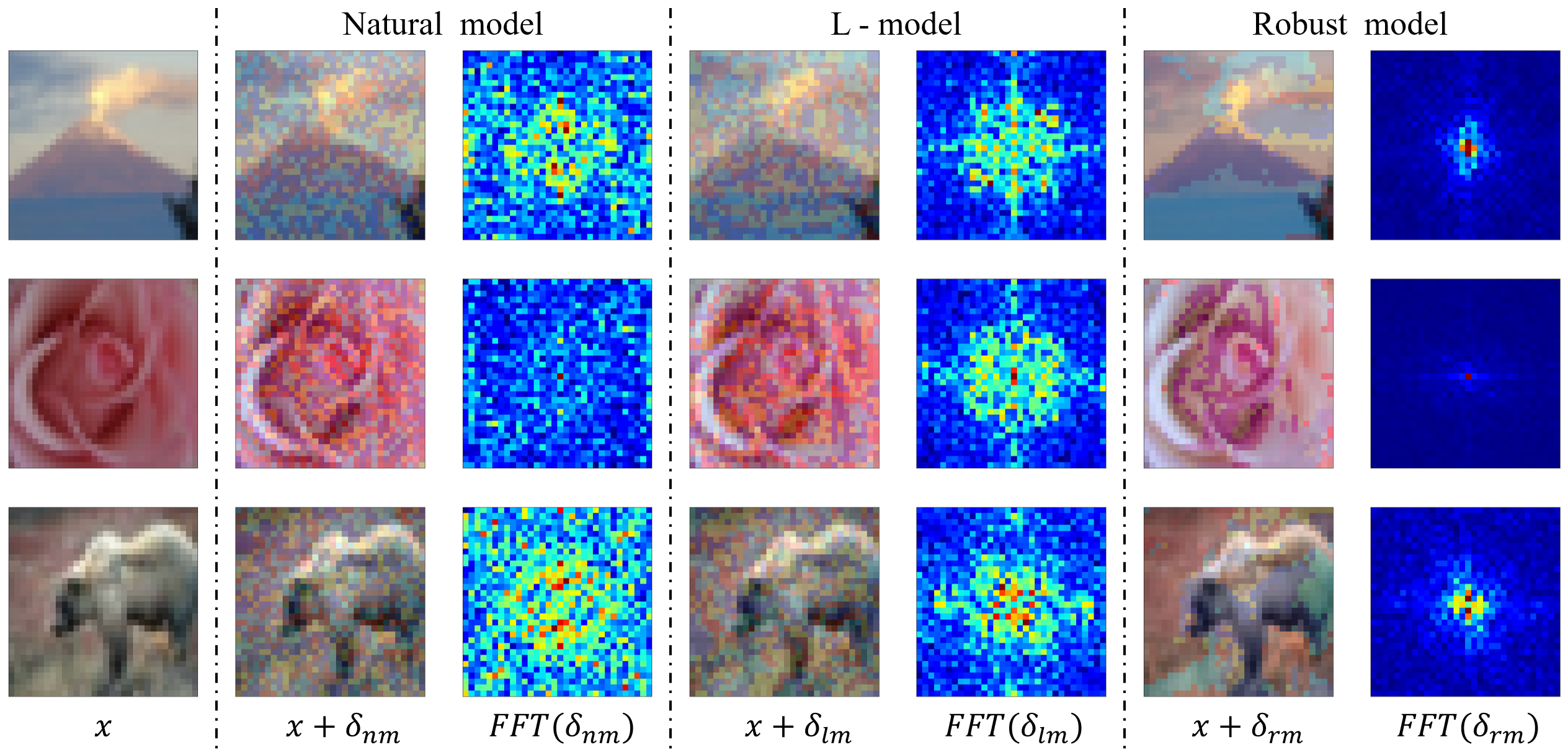}
\end{center}
\caption{Visualization of the natural and perturbed images on CIFAR-100.}
\label{c100_delta}
\end{figure}

\begin{figure}[ht]
\begin{center}
%\framebox[4.0in]{$\;$}
\includegraphics[width=0.95\linewidth]{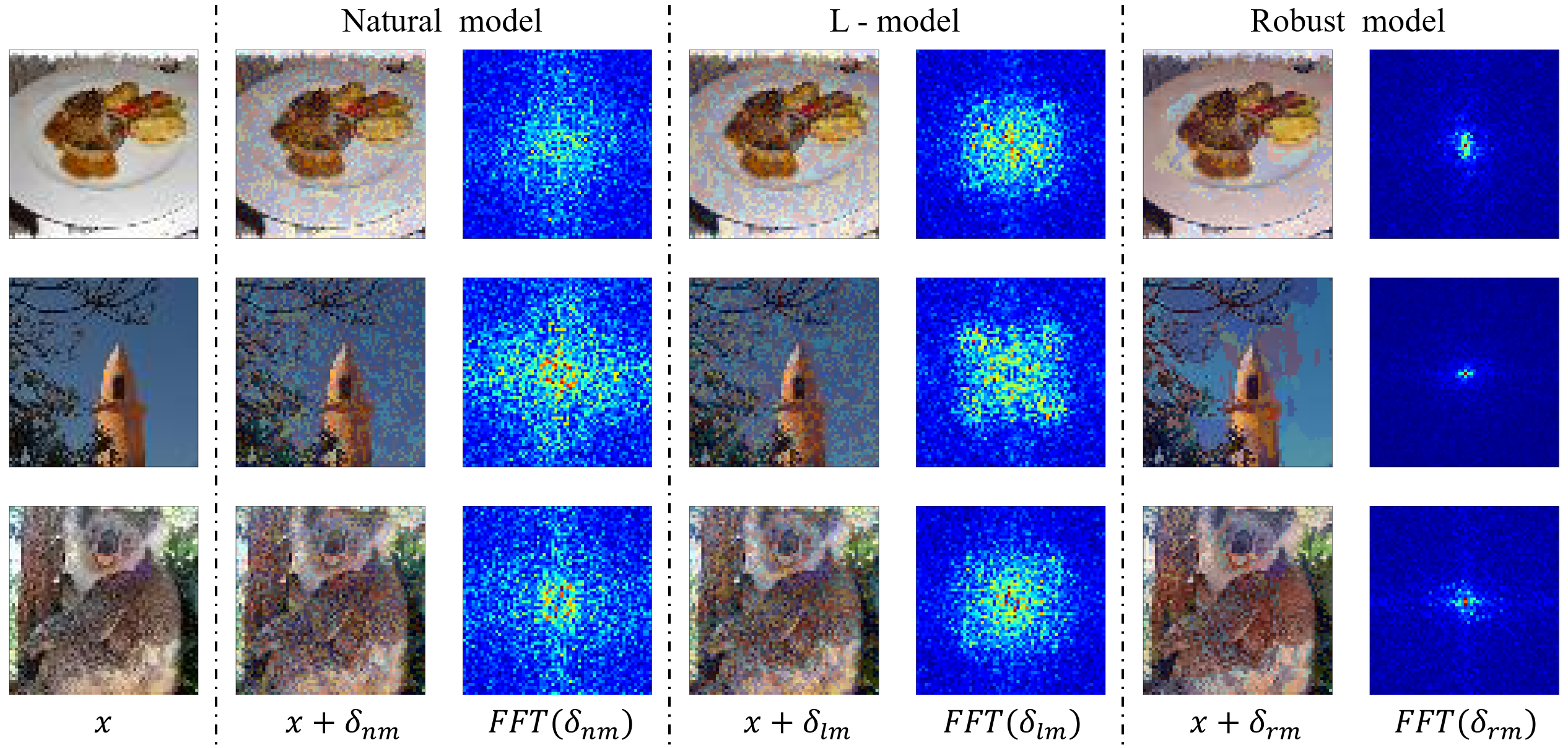}
\end{center}
\vspace{-4mm}
\caption{Visualization of the natural and perturbed images on Tiny ImageNet.}
\label{tin_delta}
\end{figure}

Figures~\ref{c100_delta}\&~\ref{tin_delta} show the natural images, adversarial images, and the spectral distribution (low frequency in the center) of the perturbations across the datasets. $x$ denotes the natural images, $\delta_{nm}$, $\delta_{lm}$ and $\delta_{rm}$ denote the PGD-20 attack perturbations generated according to the natural, L- and robust models, respectively. $FFT$ denotes the Fast Fourier Transform. \textup{Jet} color map is used to highlight perturbations for clear visualization. For the natural models, the perturbations are a jumble of noise points within the picture and have larger magnitudes in the high-frequency region than L- and robust models. The perturbations for the robust models are significantly more ordered and mainly concentrated in the low-frequency region. These visualizations prove that the adversarial perturbation is not a simple high-frequency phenomenon and is model- and dataset-dependent.

% \begin{figure}[ht]
% \begin{center}
% %\framebox[4.0in]{$\;$}
% \includegraphics[width=0.95\linewidth]{Fig/svhn.png}
% \end{center}
% \caption{Visualization of the natural and perturbed images on SVHN.}
% \label{svhn_delta}
% \end{figure}

\subsection{Detailed Settings for Popular Defenses}
\label{popular}
\paragraph{TRADES~\cite{zhang2019theoretically}} It decomposes the robust error as the sum of the natural error and the boundary error and encourages the algorithm to push the decision boundary away from the data to improve the robust accuracy. The overall loss function is shown as follows:
\begin{equation}
\label{trades}
\mathcal{L}_{AT} = \mathbf{CE}(f(x), y) + \lambda \cdot \mathbf{KL}(f(x) || f(x+\delta))
\end{equation}
$CE$ denotes the Cross-Entropy loss, $KL$ denotes the Kullback-Leibler divergence generated by PGD-10, $\delta$ denotes the adversarial PGD-10 perturbations, $f(x)$ denotes the probability predicted by the model, $y$ denotes the true label, $\lambda$ is the coefficient to balance the $CE$ and $KL$ loss. Following the default setting in TRADES, we adopt SGD with momentum 0.9, weight decay $2\times10^{-4}$, and batch size 128. The model is trained for 100 epochs on one 3090 GPU. The initial learning rate is set to 0.1 , which decays to one-tenth at 75th and 90th epochs, respectively. The $\lambda$ is set to 6.

\paragraph{MART~\cite{wang2019improving}} Based on standard AT, it explicitly differentiates the misclassified and correctly classified examples during the training and adds a misclassification-aware regularization to the standard adversarial risk to achieve better robustness. The overall loss function is shown as follows:
\begin{equation}
\label{mart}
\mathcal{L}_{AT} = \mathbf{BCE}(f(x+\delta), y) + \lambda \cdot \mathbf{KL}(f(x) || f(x+\delta))\cdot(1-f(x))
\end{equation}
$BCE$ denotes the binary Cross-Entropy loss. Following the default setting in MART, we adopt SGD with momentum 0.9, weight decay $2\times10^{-4}$, and batch size 128. The model is trained for 100 epochs on one 3090 GPU. The initial learning rate is set to 0.1 , which decays to one-tenth at 75th and 90th epochs, respectively. The $\lambda$ is set to 6.

\paragraph{AWP~\cite{wu2020adversarial}} It identifies the connection between the weight loss landscape and the robust generalization gap, proposes adversarial weight perturbation to directly make the weight loss landscape flat, and develops a double perturbation (perturbing both inputs and weights) mechanism in the AT framework.
Following the default setting in AWP, we adopt SGD with momentum 0.9, weight decay $5\times10^{-4}$, and batch size 128. The model is trained for 200 epochs on two V100 GPUs. The initial learning rate is 0.1, which decays to one-tenth at 100th and 150th epochs, respectively.

\subsection{Plug-and-play blocks across datasets}
\label{more_experiments}
We apply SAR and SARWA to popular defenses to verify the effectiveness of the plug-and-play modules on datasets CIFAR-100 and Tiny ImageNet. The results are shown in Table~\ref{c100_more} and Table ~\ref{tin_more}). 

\begin{table}[h]
\caption{Top-1 accuracy(\%) of the ResNet18 model on the CIFAR-100. Bold \textbf{\textcolor{red}{red}} and \textbf{\textcolor{blue}{blue}} numbers indicate the top and runner-up.}
\label{c100_more}
\begin{center}
\resizebox{\linewidth}{!}{
\begin{tabular}{c|ccccc}
\hline
%\multicolumn{1}{c}{Dataset}  & \multicolumn{1}{c}{PGD-20}
Method  & Clean & PGD-20 & PGD-50 & C\&W & AA 
\\ \hline 

AT    & \textbf{\textcolor{red}{54.18}} & 27.81 & 27.49 & 25.82 & 23.56\\
SAR     & 49.23 & \textbf{\textcolor{blue}{31.27}} & \textbf{\textcolor{blue}{31.20}} & \textbf{\textcolor{blue}{27.60}} & \textbf{\textcolor{red}{26.09}}\\
SARWA & \textbf{\textcolor{blue}{53.66}} & \textbf{\textcolor{red}{31.49}} & \textbf{\textcolor{red}{31.36} }& \textbf{\textcolor{red}{28.24}}  & \textbf{\textcolor{blue}{26.06}}\\

\hline 
TRADES & \textbf{\textcolor{blue}{56.24}} & 28.48 & 28.40 & 24.71 & 23.77 \\

TRADES + SAR & 55.57 & \textbf{\textcolor{blue}{30.08}} & \textbf{\textcolor{blue}{29.95}} & \textbf{\textcolor{blue}{25.93}} & \textbf{\textcolor{blue}{25.05}} \\

TRADES + SARWA & \textbf{\textcolor{red}{56.59}} & \textbf{\textcolor{red}{30.22}} & \textbf{\textcolor{red}{30.20}} & \textbf{\textcolor{red}{26.81}} & \textbf{\textcolor{red}{25.37}} \\

\hline 
MART & \textbf{\textcolor{red}{51.23}} & 29.66 & 29.55 & 25.88 & 24.27 \\

MART + SAR & 49.33 & \textbf{\textcolor{blue}{31.03}} & \textbf{\textcolor{blue}{30.87}} & \textbf{\textcolor{blue}{26.78}} & \textbf{\textcolor{blue}{25.07}} \\

MART + SARWA & \textbf{\textcolor{blue}{50.72}} & \textbf{\textcolor{red}{31.75}} & \textbf{\textcolor{red}{31.68}} & \textbf{\textcolor{red}{27.32}} & \textbf{\textcolor{red}{25.46}} \\

\hline 
AWP & \textbf{\textcolor{blue}{54.71}} & 30.88 & 30.69 & 27.87 & 25.74 \\

AWP + SAR & 48.92 & \textbf{\textcolor{red}{31.90}} & \textbf{\textcolor{red}{31.73}} & \textbf{\textcolor{blue}{28.02}} & \textbf{\textcolor{blue}{26.10}} \\

AWP + SARWA & \textbf{\textcolor{red}{55.78}} & \textbf{\textcolor{blue}{31.75}} & \textbf{\textcolor{blue}{31.60}} & \textbf{\textcolor{red}{28.84}} & \textbf{\textcolor{red}{26.64}} \\

\hline

\end{tabular}}
\end{center}
\end{table}

\begin{table}[H]
\caption{Top-1 accuracy(\%) of the ResNet18 model on the Tiny ImageNet. Bold \textbf{\textcolor{red}{red}} and \textbf{\textcolor{blue}{blue}} numbers indicate the top and runner-up.}
\label{tin_more}
% \begin{center}
% \setlength{\tabcolsep}{3.5mm}{
\resizebox{\linewidth}{!}{
\begin{tabular}{c|ccccc}
\hline
%\multicolumn{1}{c}{Dataset}  & \multicolumn{1}{c}{PGD-20}
Method  & Clean & PGD-20 & PGD-50 & C\&W & AA 
\\ \hline 
 AT    & \textbf{\textcolor{red}{46.64}} & 23.33 & 23.18 & 20.44 & 18.34\\
 SAR     & 42.92 & \textbf{\textcolor{blue}{25.32}} & \textbf{\textcolor{blue}{25.25}} & \textbf{\textcolor{blue}{21.35}} & \textbf{\textcolor{red}{19.71}} \\
 SARWA & \textbf{\textcolor{blue}{46.55}} & \textbf{\textcolor{red}{25.64}} & \textbf{\textcolor{red}{25.47}} & \textbf{\textcolor{red}{21.90}} & \textbf{\textcolor{blue}{19.64}}\\
 
\hline 
TRADES & \textbf{\textcolor{blue}{48.33}} & 22.77 & 22.71 & 18.89 & 17.95 \\

TRADES + SAR & 46.61 & \textbf{\textcolor{blue}{23.79}} & \textbf{\textcolor{blue}{23.72}} & \textbf{\textcolor{blue}{19.97}} & \textbf{\textcolor{blue}{18.99}} \\

TRADES + SARWA & \textbf{\textcolor{red}{48.94}} & \textbf{\textcolor{red}{24.81}} & \textbf{\textcolor{red}{24.73}} & \textbf{\textcolor{red}{20.21}} & \textbf{\textcolor{red}{19.34}} \\

\hline 
MART & \textbf{\textcolor{red}{45.39}} & 24.17 & 24.09 & 20.07 & 18.49 \\

MART + SAR & 43.35 & \textbf{\textcolor{blue}{25.35}} & \textbf{\textcolor{blue}{25.31}} & \textbf{\textcolor{blue}{20.63}} & \textbf{\textcolor{blue}{19.31}} \\

MART + SARWA & \textbf{\textcolor{blue}{44.79}} & \textbf{\textcolor{red}{25.83}} & \textbf{\textcolor{red}{25.80}} & \textbf{\textcolor{red}{21.53}} & \textbf{\textcolor{red}{19.44}} \\
\hline 
AWP & \textbf{\textcolor{red}{46.49}} & 24.76 & 24.59 & 21.04 & 19.11 \\

AWP + SAR & 44.93 & \textbf{\textcolor{red}{24.98}} & \textbf{\textcolor{red}{24.90}} & \textbf{\textcolor{red}{21.54}} & \textbf{\textcolor{red}{19.52}} \\

AWP + SARWA & \textbf{\textcolor{blue}{46.32}} & \textbf{\textcolor{blue}{24.86}} & \textbf{\textcolor{blue}{24.77}} & \textbf{\textcolor{blue}{21.32}} & \textbf{\textcolor{blue}{19.38}} \\

\hline

\end{tabular}}
% \end{center}
\end{table}

Experimental results demonstrate that the proposed methods can be plugged into these popular defenses to further improve robustness. Additionally, SARWA can achieve a comparable level of standard accuracy to AT, while simultaneously enhancing robust accuracy, which is a significant accomplishment.

\vfill

\end{document}